%% file: acl_latex.tex
\pdfoutput=1

\documentclass[11pt]{article}

\usepackage[preprint]{acl}

\usepackage{times}
\usepackage{latexsym}

\usepackage[T1]{fontenc}

\usepackage[utf8]{inputenc}

\usepackage{microtype}

\usepackage{inconsolata}

\usepackage{graphicx}

\usepackage{amsmath}
\usepackage{amssymb}
\usepackage{amsthm}
\usepackage{booktabs}
\usepackage{placeins}
\usepackage{colortbl}
\usepackage{bbm}
\usepackage{enumitem}

\newtheoremstyle{hypstyle}{\topsep}{\topsep}{\itshape}{0pt}%
    {\bfseries}{.}{ }{\thmname{#1}\thmnumber{ #2}\thmnote{ (#3)}}
\theoremstyle{hypstyle}
\newtheorem{hypothesis}{Hypothesis}

\newcommand{\RL}{\mathrm{RL}}
\newcommand{\Kn}{\mathrm{K}}
\newcommand{\Rs}{\mathrm{R}}
\newcommand{\ind}{\mathbbm{1}}
\newcommand{\Ex}{\mathbb{E}}

\title{The Reliance Hypothesis: Chain-of-Thought Faithfulness Rises as Language Resource Level Falls}

\author{%
    Lucas Resck\textnormal{\textsuperscript{1}},
    Jingyi Sun\textnormal{\textsuperscript{2}},
    Qianli Wang\textnormal{\textsuperscript{3,4}},
    Isabelle Augenstein\textnormal{\textsuperscript{2}},
    Anna Korhonen\textnormal{\textsuperscript{1}} \\%
    \textsuperscript{1}Language Technology Lab, University of Cambridge \quad
    \textsuperscript{2}University of Copenhagen \\%
    \textsuperscript{3}Technische Universit\"at Berlin \quad
    \textsuperscript{4}German Research Center for Artificial Intelligence (DFKI) \\%
    \small{%
        \texttt{\{ler44, alk23\}@cam.ac.uk}, \texttt{\{jisu, augenstein\}@di.ku.dk}, \texttt{qianli.wang@tu-berlin.de}%
    }%
}

\begin{document}

    \maketitle

    \begin{abstract}
        \input{content/abstract}
    \end{abstract}

    \input{content/introduction}

    \input{content/related_work}

    \input{content/cross_lingual_faithfulness}

    \input{content/reliance_hypothesis}

    \input{content/discussion}

    \newpage

    \input{content/limitations}

    \input{content/acknowledgements}



\input{acl_latex.bbl}
    \appendix

    \input{content/appendix}

\end{document}

%% file: content/abstract.tex
Chain-of-Thought (CoT) faithfulness matters for monitoring, explaining and safely deploying large language models, yet it has been studied mostly in English.
Languages differ in \emph{resource level}, the amount of text available to learn from, so models know less in lower-resource languages, which may change how much they rely on their reasoning.
To our knowledge, we are the first to relate CoT faithfulness to both resource level and task type (knowledge- or reasoning-intensive).
On parallel multiple-choice questions from Multilingual Massive Multitask Language Understanding (MMMLU) in eight languages, faithfulness rises as resource level falls on knowledge-intensive tasks but not on reasoning-intensive ones, a gap that points the same way across four models and several faithfulness measures.
We propose, and find support for, the Reliance Hypothesis: models rely more on their reasoning when they cannot recall the answer, as in lower-resource languages on knowledge-intensive tasks.
Model confidence partly explains the effect of resource level, but not the knowledge--reasoning gap.
Languages in which models are less confident rely more on reasoning, yet less confident answers within a language do not, a Simpson's paradox.
In lower-resource languages, models rely on their reasoning for more answers, not more strongly for each.
CoT monitoring may therefore be most informative in lower-resource languages, where it is least often validated.

%% file: content/introduction.tex
\section{Introduction}

    \begin{figure}[t]
        \centering
        \includegraphics[width=\columnwidth]{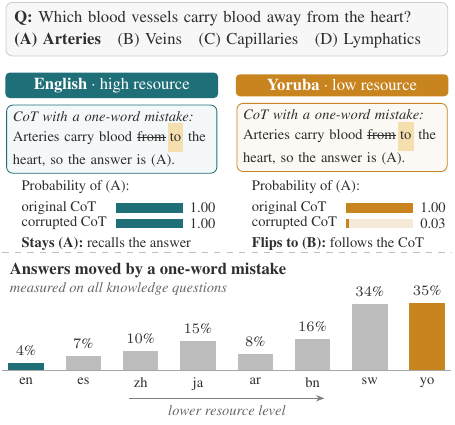}
        \caption{The \emph{Reliance Hypothesis} at a glance.
            \emph{Top}: an illustrative knowledge question whose CoT is corrupted by a one-word mistake (from $\to$ to). A model answering in a high-resource language falls back on the answer it recalls and keeps (A); in a low-resource language, where its knowledge is thinner, it follows its reasoning instead, and the probability of (A) collapses.
            \emph{Bottom}: measured on all knowledge questions, the share of answers whose probability a one-word mistake moves (\autoref{fig:rate_magnitude}).}
        \label{fig:intro}
    \end{figure}

    Chain-of-Thought (CoT) prompting elicits intermediate reasoning from large language models (LLMs) before they answer \cite{wei_chain--thought_2022}.
    Beyond improving accuracy, the CoT offers a window into how a model reaches its answer, which makes it valuable for monitoring and safety \cite{korbak_chain_2025, chen2025reasoning} and for explaining their decisions \cite{turpin2023language, parcalabescu-frank-2024-measuring}.
    This value, however, depends on CoT \emph{faithfulness}: the CoT must reflect the process by which the model actually reaches its answer \cite{jacovi-goldberg-2020-towards}.

    A growing body of work measures CoT faithfulness \cite{lanham2023measuring, turpin2023language, tutek_measuring_2025, zaman_causal_2025}, improves it \cite{paul-etal-2024-making,sun2026investigatinginterplaycontextualparametric} and probes its mechanisms \cite{macar2026thought, ye2026mechanistic, chen2026does}, but almost entirely in English \cite{ghosh2025surveymultilingualreasoninglanguage}.
    The few multilingual studies report that faithfulness differs across languages \cite{zhao_comprehensive_2026, onyame2026fragilitychainofthoughtmonitoringtypologically}, yet leave open how it relates to a language's resource level, and why.
    This matters because models reason differently across languages \cite{zhao_comprehensive_2026, ghosh2025surveymultilingualreasoninglanguage} and hold inconsistent factual knowledge across them \cite{qi_cross-lingual_2023}, so how much a model leans on its CoT may depend on the language it answers in.

    Consider the knowledge question in \autoref{fig:intro}, with one word of the reasoning corrupted.
    In English, the model can recall the answer and ignores the mistake; in Yoruba, where its knowledge is thinner, it follows the mistake to a different answer.
    We formalise this intuition as the \emph{Reliance Hypothesis}: models rely more on their reasoning when they cannot simply recall the answer, as in lower-resource languages on knowledge-intensive tasks.

    We test the hypothesis on parallel multiple-choice questions from Multilingual Massive Multitask Language Understanding (MMMLU; \citealp{openai_multilingual_2024}), which covers both knowledge- and reasoning-intensive subjects, in eight languages whose resource levels span nearly eight orders of magnitude, with four open-weight models and four interventions on the reasoning chain \cite{lanham2023measuring}.
    We measure faithfulness as reliance on the CoT: how far each intervention shifts the model's confidence in its original answer, and compare knowledge- and reasoning-intensive subtasks, so that the same languages and models serve as their own control.
    Our contributions are as follows:
    \begin{enumerate}[noitemsep]
        \item[(i)] To our knowledge, we are the first to relate CoT faithfulness to both resource level and task type.
        \item[(ii)] We show that faithfulness rises as resource level falls on knowledge-intensive tasks, but not on reasoning-intensive ones, a gap that points the same way across models and metrics and largely survives
            robustness checks.
        \item[(iii)] We propose and test the \emph{Reliance Hypothesis}, finding that model confidence partly mediates the effect of resource level, but that confidence alone cannot explain why the effect is stronger on knowledge-intensive tasks.
        \item[(iv)] We characterise reliance further: models rely more on reasoning in languages where they are less confident, yet within a language, their less confident answers rely on it no more than others (a Simpson's paradox: a trend across groups that vanishes within them); and lower-resource languages rely on their reasoning more often, not more heavily (4\% of answers in English vs. 35\% in Yoruba).\looseness=-1
    \end{enumerate}
    Faithfulness is thus highest where models know least, suggesting CoT monitoring may be most informative where it is least often validated.

%% file: content/related_work.tex
\section{Related Work}

    \paragraph{Measuring CoT Faithfulness.}
        Absent ground truth, how to measure CoT faithfulness remains unsettled.
        \emph{Contextual} approaches intervene on the input or on the chain: hint-based tests deem a chain unfaithful if an inserted cue sways the answer without being mentioned \cite{turpin2023language, chen2025reasoning, chua2025deepseekr1reasoningmodels}, while other tests truncate the chain, replace it with filler or add mistakes, and check whether the answer or the model's confidence follows \cite{lanham2023measuring, ye2026mechanistic}.
        \emph{Parametric} approaches unlearn the knowledge stated in reasoning steps \cite{tutek_measuring_2025, sun2026investigatinginterplaycontextualparametric}; others compare the chain with input attributions \cite{parcalabescu-frank-2024-measuring} or inspect internal computations \cite{shen2026detectingunfaithfulchainofthoughtcircuitguided, occhipinti2026probing, wang2026conceptalignmentcausalgrounding, chen2026does}.
        Following \citet{lanham2023measuring, zaman_causal_2025}, we adopt four interventions, covering both removal and corruption of the CoT, and three metrics (\autoref{subsec:experimental_setup}).

    \paragraph{Faithfulness Across Languages.}
        Faithfulness across languages has only recently attracted attention \cite{ghosh2025surveymultilingualreasoninglanguage}.
        In Persian, an LLM judge finds faithfulness driven more by the dataset and model than by the prompting language \cite{yazdani_comprehensive_2026}, but such a judge sees only whether a CoT appears to support the answer, not whether the model used it.
        On maths problems in 11 languages, truncating or corrupting thinking traces instead shows stronger reliance on them outside English \cite{zhao_comprehensive_2026}, while adversarial hints go unmentioned in nearly all chains \cite{onyame2026fragilitychainofthoughtmonitoringtypologically}; the two need not conflict, since a chain can drive the answer without mentioning a hint \cite{zaman-srivastava-2026-chain}.
        None of these studies relates faithfulness to resource level or contrasts knowledge- and reasoning-intensive tasks.
        From the pattern we find, we derive the \emph{Reliance Hypothesis}, which parallels evidence that models gain most from external context on facts they rarely memorise \cite{mallen_when_2023}.

%% file: content/cross_lingual_faithfulness.tex
\section{Cross-lingual Faithfulness Variation}

    This section examines how CoT faithfulness varies across languages.
    After describing our setup, we show that faithfulness rises as resource level falls on knowledge-intensive tasks but not on reasoning-intensive ones, consistently across metrics, models and subtasks.

    \input{content/cross_lingual_faithfulness/experimental_setup}

    \input{content/cross_lingual_faithfulness/faithfulness_varies}

    \input{content/cross_lingual_faithfulness/different_metrics}

    \input{content/cross_lingual_faithfulness/models_agree}

    \input{content/cross_lingual_faithfulness/knowledge_reasoning_gap}

%% file: content/cross_lingual_faithfulness/experimental_setup.tex
\subsection{Experimental Setup}
    \label{subsec:experimental_setup}

    \paragraph{CoT Faithfulness Metrics.}

        To measure faithfulness, we intervene on the reasoning chain and observe how the model's confidence in its original answer changes \cite{lanham2023measuring}.
        We use four interventions:
        (i) \emph{Early Answering} keeps only the first third of the chain;
        (ii) \emph{Filler Tokens} replaces every character of the chain with a dot;
        (iii) \emph{Adding Mistakes} replaces one word in the chain with its antonym; and
        (iv) \emph{Paraphrasing} rewrites the first three quarters of the chain and lets the model complete the rest.%
        \footnote{\autoref{app:cot_interventions} describes the interventions in detail.}
        The first two \emph{remove} reasoning; the last two \emph{corrupt} it.
        Although \citet{lanham2023measuring} intend paraphrases to preserve meaning, ours alter the chain substantially and change the answer twice as often as Adding Mistakes, so we treat Paraphrasing as corrupting (\autoref{app:cot_interventions}).

        Let $z = P(a \mid x; c)$ be the probability that the model assigns to its answer $a$ given the input $x$ and its CoT $c$, and let $z' = P(a \mid x; c')$ be the probability after an intervention.
        We report three metrics:\looseness=-1
        \begin{align}
            \Delta p &= z - z', \label{eq:raw} \\
            \Delta \log p &= \log z - \log z', \label{eq:delta_log} \\
            \Phi &= \ind\left[\arg\max_v P(v \mid x; c') \neq a\right], \label{eq:flip}
        \end{align}
        where $v$ ranges over the answer options.
        We mainly use $\Delta \log p$, the relative drop.
        Unlike $\Delta p$, it is not capped by $z$, which is lower in low-resource languages (\autoref{app:data_sanity}), yet it ranks samples almost identically (\autoref{sec:different_metrics}); unlike $\Phi$, it also registers drops that leave the answer unchanged.

    \paragraph{Data \& Languages.}

        \begin{table}[t]
            \centering
            \small
            \setlength{\tabcolsep}{4pt}
            \input{tables/tab_languages}\hspace{1em}
            \input{tables/tab_subtasks}
            \caption{Evaluation languages (left) and MMMLU subtasks, grouped by
                type (right). $\RL_\ell$ is $\log_{10}$ of the language's CulturaX
                token count. \emph{Items}: questions per subtask, the same ones in every language.}
            \label{tab:languages-subtasks}
        \end{table}

        For English, we use MMLU \cite{hendrycks_measuring_2021}; for other languages, we use MMMLU~\cite{openai_multilingual_2024}, its professional human translation, so all languages share the same questions.
        We select eight of MMMLU's 15 languages, from five families and five scripts, and eight subtasks, half knowledge-intensive and half reasoning-intensive, with up to 250 questions each (\autoref{tab:languages-subtasks}).
        The languages' resource levels, measured by token counts in the Common Crawl corpus CulturaX~\cite{nguyen_culturax_2024}, a common proxy for pretraining data \cite{lai_okapi_2023, resck_explainability_2025, resck_mitigating_2026}, span nearly eight orders of magnitude.
        \autoref{app:data_sanity} analyses the model-generated CoTs and answers in each language.

    \paragraph{Models.}

        We evaluate four instruction-tuned, multilingual LLMs: \texttt{Gemma-3-4B-it}~\cite{gemma_team_gemma_2025}, \texttt{Qwen2.5-7B-Instruct}~\cite{qwen_qwen25_2025}, \texttt{Llama-3.1-8B-Instruct}~\cite{grattafiori_llama_2024} and \texttt{Aya-Expanse-8B}~\cite{dang_aya_2024}.
        We use the larger \texttt{Qwen2.5-14B-Instruct} as the \emph{perturber} to edit the chain for Adding Mistakes and Paraphrasing.
        All models are open-weight, giving access to the required token log-probabilities.

%% file: tables/tab_languages.tex
\begin{tabular}[t]{@{}lr@{}}
\toprule
\textbf{Language} & $\boldsymbol{\mathrm{RL}_\ell}$ \\
\midrule
English & $12.45$ \\
Spanish & $11.57$ \\
Chinese & $11.36$ \\
Japanese & $11.03$ \\
Arabic & $10.84$ \\
Bengali & $9.98$ \\
Swahili & $7.49$ \\
Yoruba & $4.63$ \\
\bottomrule
\end{tabular}

%% file: tables/tab_subtasks.tex
\begin{tabular}[t]{@{}lr@{}}
\toprule
\textbf{Subtask} & \textbf{Items} \\
\midrule
\multicolumn{2}{@{}l@{}}{\emph{Knowledge-intensive}} \\
\quad Clinical Knowledge & $250$ \\
\quad Professional Medicine & $250$ \\
\quad World Religions & $171$ \\
\quad Global Facts & $100$ \\
\addlinespace
\multicolumn{2}{@{}l@{}}{\emph{Reasoning-intensive}} \\
\quad High School Mathematics & $250$ \\
\quad Formal Logic & $126$ \\
\quad Abstract Algebra & $100$ \\
\quad College Mathematics & $100$ \\
\addlinespace
\quad \emph{Total} & $1{,}347$ \\
\bottomrule
\end{tabular}

%% file: content/cross_lingual_faithfulness/faithfulness_varies.tex
\subsection{Faithfulness Varies Across Languages}
    \label{sec:faithfulness_varies}

    \begin{figure*}[t]
        \centering
        \includegraphics[width=\textwidth]{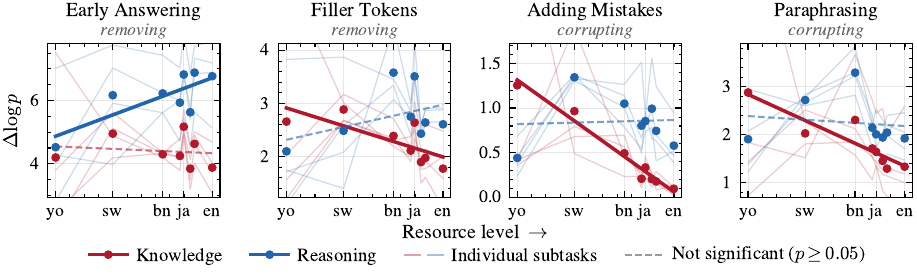}

        \vspace{6pt}
        \begin{minipage}[c]{0.38\textwidth}
            \centering
            \small
            \setlength{\tabcolsep}{5pt}
            \input{tables/tab_slopes_dlogp}
        \end{minipage}\hfill
        \begin{minipage}[c]{0.59\textwidth}
            \caption{Faithfulness ($\Delta \log p$) against language resource level.
                \emph{Top}: mean faithfulness per language on knowledge- and reasoning-intensive subtasks (markers), with slopes from \autoref{eq:slope} (lines); faint lines show individual subtasks, clipped to the axes.
                \emph{Left}: the slopes for knowledge ($\hat\beta^{\mathrm{K}}$) and reasoning ($\hat\beta^{\mathrm{R}}$) subtasks, per order of magnitude of token count, and their difference.
                Solid lines and bold values mark significant slopes ($p<.05$).}
            \label{fig:variation}
        \end{minipage}
    \end{figure*}

    To quantify how faithfulness varies across languages, we fit a linear regression with fixed effects:
    \begin{equation}
        \bar{y}_{ms\ell} = \alpha_m + \gamma_s + \beta\,\RL_\ell + \varepsilon_{ms\ell},
        \label{eq:slope}
    \end{equation}
    where $\bar{y}_{ms\ell}$ is the mean faithfulness (Eqs.~\eqref{eq:raw}--\eqref{eq:flip}) of model $m$ on subtask $s$ in language $\ell$, $\alpha_m$ and $\gamma_s$ are model and subtask fixed effects, and $\beta$ is the change in faithfulness per unit of resource level, that is, per order of magnitude of the language's token count.
    We let $\beta$ differ between knowledge- and reasoning-intensive subtasks, giving the slopes $\hat\beta^{\mathrm{K}}$ and $\hat\beta^{\mathrm{R}}$.
    Averaging within each (model, subtask, language) cell weights every model and subtask equally in every language, although the share of valid generations falls from 97\% in English to 35\% in Yoruba (\autoref{app:data_sanity}).
    Because resource level varies only between languages, the evidence for $\beta$ rests on eight languages, however many generations lie behind each cell.
    We therefore use a Fisher randomisation test~\cite{young_channeling_2019}: the $p$-value is the share of all $8! = 40{,}320$ assignments of resource levels to languages that yield a slope at least as extreme as the observed one. \looseness=-1

    \autoref{fig:variation} shows that,
    \textbf{on knowledge-intensive tasks, faithfulness rises as resource level falls} under every intervention except Early Answering.
    Under Adding Mistakes, for instance, $\hat\beta^{\mathrm{K}} = -0.16$ predicts a drop in log-probability 1.3 larger in Yoruba than in English, close to the observed gap of 1.2.
    On reasoning-intensive tasks, by contrast, no slope is significantly negative, and under Early Answering the trend even reverses.
    CoT length is a natural suspect for this exception, since the same content takes more tokens in lower-resource languages \cite{ahia_all_2023}, but controlling for it leaves the Early Answering slopes unchanged (\autoref{tab:robustness-all}).
    The difference $\hat\beta^{\mathrm{K}} - \hat\beta^{\mathrm{R}}$ is negative for all four interventions and significant for three: \textbf{lower resource levels raise faithfulness more on knowledge tasks than on reasoning tasks}.
    The other variation metrics broadly point the same way (\autoref{app:full_results_variation_metrics}).

%% file: tables/tab_slopes_dlogp.tex
\begin{tabular}{lrrr}
\toprule
\textbf{Intervention} & $\boldsymbol{\hat\beta^{\mathrm{K}}}$ & $\boldsymbol{\hat\beta^{\mathrm{R}}}$ & $\boldsymbol{\hat\beta^{\mathrm{K}} - \hat\beta^{\mathrm{R}}}$ \\
\midrule
Early & \cellcolor[rgb]{0.946,0.874,0.882}-0.08 & \cellcolor[rgb]{0.605,0.742,0.667}\textbf{\phantom{-}0.24} & \cellcolor[rgb]{0.780,0.483,0.516}\textbf{-0.31} \\
Filler & \cellcolor[rgb]{0.897,0.757,0.773}\textbf{-0.15} & \cellcolor[rgb]{0.858,0.907,0.881}\phantom{-}0.08 & \cellcolor[rgb]{0.837,0.617,0.641}\textbf{-0.23} \\
Mistakes & \cellcolor[rgb]{0.886,0.731,0.748}\textbf{-0.16} & \cellcolor[rgb]{0.990,0.994,0.992}\phantom{-}0.01 & \cellcolor[rgb]{0.882,0.722,0.739}\textbf{-0.17} \\
Paraphrasing & \cellcolor[rgb]{0.845,0.635,0.659}\textbf{-0.22} & \cellcolor[rgb]{0.981,0.956,0.959}-0.03 & \cellcolor[rgb]{0.864,0.679,0.700}-0.19 \\
\bottomrule
\end{tabular}

%% file: content/cross_lingual_faithfulness/different_metrics.tex
\subsection{Different Metrics, Same Story}
    \label{sec:different_metrics}

    Our conclusions should not hinge on the choice of variation metric, so we check how closely the three metrics agree across individual generations.
    They agree well: for every intervention, all pairwise Pearson correlations are at least 0.67, and $\Delta p$ and $\Delta \log p$ rank generations almost identically, with Spearman correlations of 0.92--1.00 (\autoref{tab:metric-corr}).
    \textbf{The two continuous metrics thus tell the same story}, even though only $\Delta p$ is capped by $z$ (\autoref{fig:metric_by_flip}).

%% file: content/cross_lingual_faithfulness/models_agree.tex
\subsection{Models Agree on Faithfulness Variation}
    \label{sec:models_agree}

    \begin{table}[t]
        \centering
        \small
        \setlength{\tabcolsep}{4pt}
        \input{tables/tab_model_agreement}
        \caption{Model agreement and each model's knowledge--reasoning gap, measured by $\Delta \log p$.
            \textbf{Agreement}: Pearson correlation between two models' mean faithfulness over languages and subtasks, computed per intervention and then averaged.
            \textbf{Gap}: mean Spearman correlation between faithfulness and resource level on knowledge (K) and reasoning (R) subtasks, averaged over subtasks and interventions, and their difference.}
        \label{tab:model_agreement}
    \end{table}

    We next ask whether our findings hold across models, in two ways (\autoref{tab:model_agreement}).
    First, for each pair of models, we correlate their mean faithfulness ($\Delta \log p$) over languages and subtasks separately by intervention,
    so that scale differences cannot drive it,
    and then average the correlations.
    Every pair agrees positively (0.10--0.52): the models broadly concur on which languages and subtasks elicit more faithful reasoning.
    This agreement is moderate, however, and weakest for Aya (0.10--0.22).

    Second, we check whether the knowledge--reasoning gap appears within each model, averaging the Spearman correlation between faithfulness and resource level
    across interventions for each subtask type.
    The correlation is lower on knowledge than on reasoning subtasks for every model, so \textbf{all four models share the gap's direction}, although its size varies from $-0.56$ (Aya) to $-0.09$ (Qwen).
    The other variation metrics tell the same story, with even stronger agreement between models (0.28--0.66), except that Qwen's gap vanishes under the answer flip $\Phi$ (\autoref{tab:model-agreement-all}).

%% file: tables/tab_model_agreement.tex
\begin{tabular}{lrrrrrr}
\toprule
& \multicolumn{3}{c}{\textbf{Agreement} ($r$)} & \multicolumn{3}{c}{\textbf{Gap} ($\bar\rho$)} \\ \cmidrule(lr){2-4}\cmidrule(l){5-7}
\textbf{Model} & \textbf{Aya} & \textbf{Gemma} & \textbf{Llama} & \textbf{K} & \textbf{R} & \textbf{K}$\,-\,$\textbf{R} \\
\midrule
Aya & -- & -- & -- & \cellcolor[rgb]{0.893,0.748,0.764}-0.49 & \cellcolor[rgb]{0.961,0.975,0.967}\phantom{-}0.07 & \cellcolor[rgb]{0.876,0.709,0.728}-0.56 \\
Gemma & \cellcolor[rgb]{0.944,0.963,0.953}0.11 & -- & -- & \cellcolor[rgb]{0.887,0.735,0.752}-0.51 & \cellcolor[rgb]{0.933,0.842,0.852}-0.31 & \cellcolor[rgb]{0.954,0.892,0.899}-0.21 \\
Llama & \cellcolor[rgb]{0.884,0.924,0.902}0.22 & \cellcolor[rgb]{0.745,0.834,0.786}0.49 & -- & \cellcolor[rgb]{0.921,0.814,0.826}-0.36 & \cellcolor[rgb]{0.994,0.996,0.995}\phantom{-}0.01 & \cellcolor[rgb]{0.918,0.808,0.820}-0.37 \\
Qwen & \cellcolor[rgb]{0.948,0.966,0.956}0.10 & \cellcolor[rgb]{0.809,0.875,0.839}0.37 & \cellcolor[rgb]{0.729,0.823,0.772}0.52 & \cellcolor[rgb]{0.980,0.987,0.983}\phantom{-}0.04 & \cellcolor[rgb]{0.932,0.956,0.943}\phantom{-}0.13 & \cellcolor[rgb]{0.980,0.953,0.956}-0.09 \\
\bottomrule
\end{tabular}

%% file: content/cross_lingual_faithfulness/knowledge_reasoning_gap.tex
\subsection{A Knowledge--Reasoning Gap}

    \begin{figure}[t]
        \centering
        \includegraphics[width=\columnwidth]{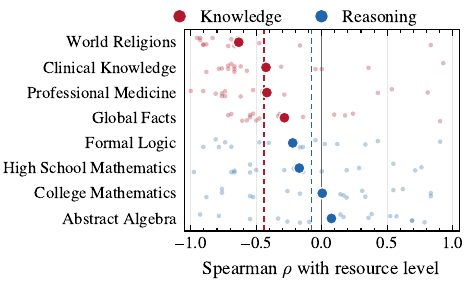}
        \caption{Trend of faithfulness ($\Delta p$) with resource level, by subtask.
            Large markers: mean Spearman correlation between faithfulness and resource level over models and interventions; small dots: individual (model, intervention) correlations; dashed lines: family means.}
        \label{fig:subtask_ranking}
    \end{figure}

    \S\S\ref{sec:faithfulness_varies} and~\ref{sec:models_agree} showed
    cross-lingual faithfulness variation that is stronger on knowledge-intensive tasks and consistent across models.
    Individual subtasks reveal the same split, motivating the Reliance Hypothesis (\autoref{sec:reliance_hypothesis}).

    \autoref{fig:subtask_ranking} ranks the subtasks by their mean correlation between faithfulness and resource level.
    \textbf{The ranking recovers the two families}: all four knowledge-intensive subtasks come first.
    Their correlations are mostly negative (75--94\% of model--intervention pairs), so faithfulness rises as resource level falls.
    Reasoning-intensive subtasks show weaker, less consistent trends, averaging $-0.22$ to $0.07$.
    We plot $\Delta p$, which gives the cleanest split; other metrics rank the subtasks similarly (\autoref{fig:subtask-ranking-appendix}).\looseness=-1

    The contrast also holds for individual answers.
    Among items a model answered validly in both English and Yoruba, Adding Mistakes changes the answer in Yoruba alone about nine times as often as in English alone on knowledge subtasks (80 vs.\ 9), but less than twice as often on reasoning subtasks (73 vs.\ 40).
    Every knowledge subtask exceeds every reasoning subtask on this ratio.
    World Religions, the steepest subtask and a largely culture-bound one, illustrates the mechanism: as accuracy falls towards Yoruba, its answers depend more on the reasoning chain (\autoref{app:world_religions}).

%% file: content/reliance_hypothesis.tex
\section{The Reliance Hypothesis}
    \label{sec:reliance_hypothesis}

    We now formalise the common cause behind these findings, the \emph{Reliance Hypothesis}: a model relies more on its reasoning when it has less, or can access less, relevant parametric knowledge for the queried fact, and this reliance drives cross-lingual variation in CoT faithfulness.
    We test two consequences: (i) the variation is stronger on knowledge-intensive than on reasoning-intensive tasks, and (ii) it is mediated by the model's confidence in its answer and reasoning chain.

    Let $\mathcal{F}$ be CoT faithfulness under Eqs.~\eqref{eq:raw}--\eqref{eq:flip}, and $\beta^{\Kn}, \beta^{\Rs}$ its resource-level slopes on knowledge- and reasoning-intensive subtasks~(\autoref{sec:faithfulness_varies}).

    \begin{hypothesis}[knowledge versus reasoning]
        \label{hyp:gap}
        When the answer can be recalled, less parametric knowledge forces the model to route it through its reasoning; when it must be computed, reasoning is needed in every language.
        Hence
        \begin{equation*}
            \beta^{\Kn} \;=\; \frac{\partial \mathcal{F}}{\partial \RL}\bigg|_{\Kn}
            \;<\; \alpha
            \;\leq\; \frac{\partial \mathcal{F}}{\partial \RL}\bigg|_{\Rs}
            \;=\; \beta^{\Rs},
        \end{equation*}
        with $\alpha \approx 0$; in particular, $\beta^{\Kn} - \beta^{\Rs} < 0$.
    \end{hypothesis}

    \begin{hypothesis}[confidence as the channel]
        \label{hyp:mediation}
        The knowledge deficit affects faithfulness \emph{through} model confidence $C$ in its answer or reasoning chain:\looseness=-1
        \begin{equation*}
            \RL \;\longrightarrow\; \text{parametric knowledge}
                \;\longrightarrow\; C \;\longrightarrow\; \mathcal{F}.
        \end{equation*}
        If $\beta_1$ is the slope in \autoref{eq:slope} and $\beta_2$ the same slope once $C$ enters the regression, this chain predicts $\lvert \beta_2 \rvert < \lvert \beta_1 \rvert$.
    \end{hypothesis}

    \input{content/reliance_hypothesis/testing}

    \input{content/reliance_hypothesis/does_confidence_explain}

    \input{content/reliance_hypothesis/simpsons_paradox}

    \input{content/reliance_hypothesis/reliance_is_rate}

    \input{content/reliance_hypothesis/robustness_tests}

%% file: content/reliance_hypothesis/testing.tex
\subsection{Does the Hypothesis Hold?}

    \begin{table}[t]
        \centering
        \small
        \input{tables/tab_hypothesis_diff}
        \caption{Test of \autoref{hyp:gap}: knowledge and reasoning slope difference, $\hat\beta^{\mathrm{K}} - \hat\beta^{\mathrm{R}}$ (\autoref{eq:twoslope}), by intervention and metric.
            Negative values support the hypothesis; \textbf{bold} indicates $p<.05$.
            Colours scale within columns.
        }
        \label{tab:hypothesis}
    \end{table}

    We first test the central prediction of \autoref{hyp:gap}, $\beta^{\Kn} - \beta^{\Rs} < 0$.
    \autoref{sec:faithfulness_varies} reported this difference under $\Delta \log p$; here we derive it from a single regression and test it under all three metrics.
    We estimate both slopes of \autoref{eq:slope} in a single fit, letting the coefficient on $\RL_\ell$ depend on whether subtask $s$ is knowledge- ($s \in \Kn$) or reasoning-intensive ($s \in \Rs$):
    \begin{align}
        \bar{y}_{ms\ell} &= \alpha_m + \gamma_s + \varepsilon_{ms\ell} \nonumber \\
                         &\quad + \big(\beta^{\Kn}\ind[s \in \Kn] + \beta^{\Rs}\ind[s \in \Rs]\big)\RL_\ell \label{eq:twoslope} \\
                         &= \alpha_m + \gamma_s + \varepsilon_{ms\ell} \nonumber \\
                         &\quad + \beta^{\Kn}\RL_\ell + (\beta^{\Rs} - \beta^{\Kn})\,\ind[s \in \Rs]\,\RL_\ell. \nonumber
    \end{align}
    The second form is the familiar interaction model: the coefficient on the interaction term $\ind[s \in \Rs]\RL_\ell$ is $\beta^{\Rs} - \beta^{\Kn}$, the negative of the difference we test and report.
    Both slopes are re-estimated under the same $8!$ relabellings of the languages, so the difference receives its own exact $p$-value.

    \autoref{tab:hypothesis} reports the difference for each intervention and metric.
    It is negative in eleven of the twelve combinations and significant in seven, including every metric under Filler Tokens and Adding Mistakes.
    Under $\Delta \log p$, it ranges from $-0.17$ to $-0.31$ per order of magnitude of token count.
    The separate slopes (\autoref{fig:variation}; \autoref{tab:slopes-all}) show the full ordering $\beta^{\Kn} < \alpha \leq \beta^{\Rs}$, with $\alpha \approx 0$, most clearly under Adding Mistakes: the knowledge slope is significantly negative under every metric, whereas the reasoning slope stays near zero.
    Paraphrasing is the exception: under $\Delta p$ and $\Phi$, faithfulness rises as resource level falls in both families, leaving little difference between them.\looseness=-1

    The answer to our question is therefore yes: \textbf{lower resource levels raise faithfulness more on knowledge-intensive than on reasoning-intensive tasks}, most clearly under Filler Tokens and Adding Mistakes.

%% file: tables/tab_hypothesis_diff.tex
\begin{tabular}{lrrr}
\toprule
\textbf{Intervention} & $\boldsymbol{\Delta p}$ & $\boldsymbol{\Delta\log p}$ & $\boldsymbol{\Phi}$ \\
\midrule
Early & \cellcolor[rgb]{0.942,0.864,0.873}-0.009 & \cellcolor[rgb]{0.780,0.483,0.516}\textbf{-0.31} & \cellcolor[rgb]{0.985,0.964,0.967}-0.002 \\
Filler & \cellcolor[rgb]{0.780,0.483,0.516}\textbf{-0.034} & \cellcolor[rgb]{0.837,0.617,0.641}\textbf{-0.23} & \cellcolor[rgb]{0.780,0.483,0.516}\textbf{-0.026} \\
Mistakes & \cellcolor[rgb]{0.896,0.754,0.770}\textbf{-0.016} & \cellcolor[rgb]{0.882,0.722,0.739}\textbf{-0.17} & \cellcolor[rgb]{0.903,0.771,0.786}\textbf{-0.011} \\
Paraphrasing & \cellcolor[rgb]{0.988,0.971,0.973}-0.002 & \cellcolor[rgb]{0.864,0.679,0.700}-0.19 & \cellcolor[rgb]{0.947,0.965,0.955}\phantom{-}0.003 \\
\bottomrule
\end{tabular}

%% file: content/reliance_hypothesis/does_confidence_explain.tex
\subsection{Does Confidence Explain the Variation?}
    \label{sec:does_confidence_explain}

    \begin{table}[t]
        \centering
        \small
        \input{tables/tab_mediation_main}
        \caption{Attenuation (\%) of the standardised resource-level slope $\beta_1$ when each confidence measure enters the regression (\autoref{eq:mediation}), for the largest significant knowledge (K) and reasoning (R) slopes under $\Delta \log p$.
            Final-answer measures are computed over the answer tokens; reasoning-chain measures over the whole chain.}
        \label{tab:mediation}
    \end{table}

    We now test \autoref{hyp:mediation}: resource level shapes faithfulness through the model's confidence $C$, either over the answer tokens (\emph{final-answer} confidence) or over the whole chain (\emph{reasoning-chain} confidence).
    For each subtask family, we compare the resource-level slope with and without $C$:
    \begin{equation}
        \beta_1 : \; \mathcal{F} \sim \RL
        \quad\text{against}\quad
        \beta_2 : \; \mathcal{F} \sim \RL + C,
        \label{eq:mediation}
    \end{equation}
    both with model and subtask fixed effects, and report the attenuation $(\beta_1 - \beta_2)/\beta_1$, the share of the slope that $C$ accounts for.
    We standardise all variables, so that slopes are comparable across metrics.

    \autoref{tab:mediation} shows the attenuation for the largest significant slope per family under $\Delta \log p$: Adding Mistakes for knowledge and Early Answering for reasoning.
    On knowledge subtasks, final-answer probability and log-probability absorb most of the slope (87\% and 70\%), whereas no reasoning-chain measure exceeds 5\%.
    On reasoning subtasks, the pattern reverses, if more weakly: reasoning-chain entropy absorbs the most (19\%), while the probability of the final answer absorbs none ($-13\%$).
    Since confidence and resource level move closely together across only eight languages, the attenuations themselves are imprecise; the robust signal is the contrast between the two kinds of confidence.
    This contrast holds in 11 of the 13 significant slopes across metrics (\autoref{tab:mediation-all}).\looseness=-1

    \emph{Can confidence alone explain the gap?}
    It cannot.
    Final-answer confidence rises with resource level at almost the same rate on knowledge and reasoning subtasks, and reasoning-chain confidence barely changes with it in either family (\autoref{tab:confidence-gradient}).
    A confidence gradient shared by both families cannot produce a faithfulness gradient that differs between them.
    Hence, \textbf{
        confidence partly explains the variation within each family, but not their gap.
    }

%% file: tables/tab_mediation_main.tex
\begin{tabular}{lrr}
\toprule
& \textbf{K}\,$\cdot$\,\textbf{Mistakes} & \textbf{R}\,$\cdot$\,\textbf{Early} \\
& $\beta_1 = -0.42$ & $\beta_1 = 0.21$ \\
\midrule
\multicolumn{3}{l}{\emph{Final answer}} \\
\quad Probability & \cellcolor[rgb]{0.543,0.702,0.615}$87$ & \cellcolor[rgb]{0.972,0.933,0.938}$-13$ \\
\quad Log-probability & \cellcolor[rgb]{0.636,0.762,0.693}$70$ & \cellcolor[rgb]{0.968,0.924,0.929}$-15$ \\
\quad Entropy & \cellcolor[rgb]{0.967,0.923,0.928}$-15$ & \cellcolor[rgb]{0.929,0.954,0.941}$14$ \\
\quad Top-1 probability & \cellcolor[rgb]{0.974,0.939,0.942}$-12$ & \cellcolor[rgb]{0.930,0.954,0.941}$13$ \\
\quad Margin & \cellcolor[rgb]{0.985,0.965,0.967}$-7$ & \cellcolor[rgb]{0.935,0.957,0.945}$13$ \\
\addlinespace
\multicolumn{3}{l}{\emph{Reasoning chain}} \\
\quad Log-probability & \cellcolor[rgb]{0.991,0.994,0.993}$2$ & \cellcolor[rgb]{0.952,0.969,0.959}$9$ \\
\quad Entropy & \cellcolor[rgb]{0.975,0.984,0.979}$5$ & \cellcolor[rgb]{0.899,0.934,0.915}$19$ \\
\quad Top-1 probability & \cellcolor[rgb]{0.992,0.982,0.983}$-3$ & \cellcolor[rgb]{0.933,0.956,0.943}$13$ \\
\quad Margin & \cellcolor[rgb]{0.976,0.944,0.948}$-11$ & \cellcolor[rgb]{0.965,0.977,0.971}$7$ \\
\bottomrule
\end{tabular}

%% file: content/reliance_hypothesis/simpsons_paradox.tex
\subsection{Is It a Simpson's Paradox?}

    \begin{figure}[t]
        \centering
        \includegraphics[width=\columnwidth]{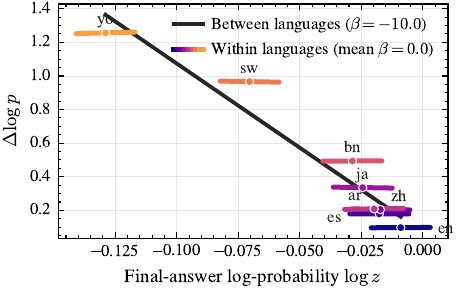}
        \caption{Faithfulness ($\Delta \log p$) against final-answer log-probability on knowledge-intensive subtasks under Adding Mistakes.
            Markers are language means, and the solid line is the fit through them; each short segment shows the slope of the fit \emph{within} one language, estimated on all of its items and drawn through its mean.}
        \label{fig:simpson}
    \end{figure}

    \autoref{hyp:mediation} posits a chain from resource level to faithfulness through confidence:
    \begin{equation*}
        \RL \;\overset{(1)}{\longrightarrow}\; C \;\overset{(2)}{\longrightarrow}\; \mathcal{F}.
    \end{equation*}
    \autoref{sec:does_confidence_explain} examined link~(1): confidence rises with resource level at the same rate in both families, and partly mediates its effect on faithfulness.
    We now examine link~(2): is a less confident answer itself more reliant on its reasoning, or does the association exist only between languages?

    \autoref{fig:simpson} shows faithfulness ($\Delta \log p$) against final-answer log-probability on knowledge-intensive subtasks under Adding Mistakes, the setting where confidence mediates the most (\autoref{tab:mediation}).
    Between languages, the association is strongly negative ($\beta = -10.0$, $r = -0.97$); within each language, it vanishes (mean $\beta \approx 0$, $r = -0.01$).
    This is a structural Simpson's paradox: a trend that appears across groups disappears, or reverses, within each group, here each language.
    The pattern is general (\autoref{tab:simpson-summary}): across all 216 combinations of metric, intervention, confidence measure and family, the mean absolute correlation is 0.47 between languages but 0.06 within them, and the within-language correlation stays below 0.1 in absolute value in 76\% of cases.

    The paradox, however, involves no contradiction.
    Within languages, the association mostly disappears rather than reverses; and a language-level cause, such as the knowledge deficit that the Reliance Hypothesis posits, should produce exactly this: an association between languages and none between individual items.
    Hence, \textbf{the link between confidence and reliance is a property of languages, not of individually uncertain items}.

%% file: content/reliance_hypothesis/reliance_is_rate.tex
\subsection{Reliance is Rate, not Magnitude}

    \begin{figure*}[t]
        \centering
        \includegraphics[width=\textwidth]{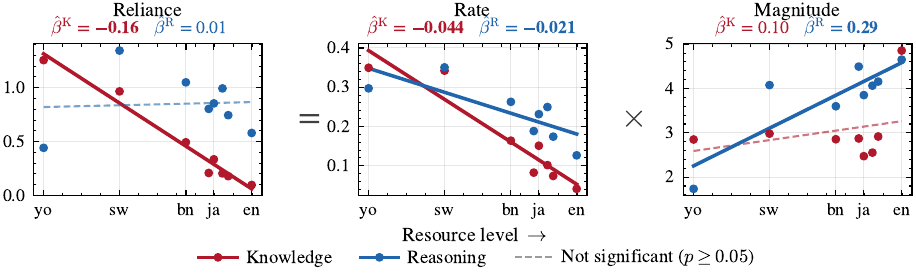}
        \caption{$\text{Reliance} = \text{Rate} \times \text{Magnitude}$ (\autoref{eq:decomp}) against resource level, under Adding Mistakes, with the slopes of \autoref{eq:slope} on knowledge and reasoning subtasks under each title (\textbf{bold}: $p<.05$); line style as in \autoref{fig:variation}.}
        \label{fig:rate_magnitude}
    \end{figure*}

    The previous sections established that reliance on the reasoning chain varies across languages, but not how: do models in lower-resource languages rely on it more \emph{often}, or more \emph{heavily}?
    Taking $\mathcal{F} = \Delta \log p$, reliance separates the two:
    \begin{equation}
        \underbrace{\Ex\!\left[\mathcal{F}\right]}_{\text{reliance}}
        = \underbrace{P\!\left(\mathcal{F} > \tau\right)}_{\text{rate}}
        \times \underbrace{\Ex\!\left[\mathcal{F} \,\middle|\, \mathcal{F} > \tau\right]}_{\text{magnitude}},
        \label{eq:decomp}
    \end{equation}
    where a small threshold, $\tau = 0.01$, absorbs numerical noise.%
    \footnote{Strictly, rate $\times$ magnitude omits $\Ex\!\left[\mathcal{F}\,\ind[\mathcal{F} \leq \tau]\right]$, which is negligible: values in $(0, \tau]$ are at most $\tau$, and since $z' \leq 1$, negative values are bounded below by $\log z$, which is close to zero because the model's own answer is usually highly probable (median $z = 1.00$). Empirically, the omitted term changes the mean by less than 1\%.}
    The rate is the chance that the intervention moves the answer probability at all; the magnitude is the average size of that move.

    \autoref{fig:rate_magnitude} plots reliance and its two factors against resource level under Adding Mistakes.
    On knowledge-intensive subtasks, the whole gradient sits in the rate: the intervention moves the answer far more often in lower-resource languages ($\hat\beta^{\Kn} = -0.044$), but not by more when it does ($\hat\beta^{\Kn} = 0.10$, not significant).
    The rate climbs from 4\% of items in English to 35\% in Yoruba, more than eightfold.
    On reasoning-intensive subtasks, the rate also falls with resource level ($\hat\beta^{\Rs} = -0.021$), but the magnitude rises ($\hat\beta^{\Rs} = 0.29$); the two cancel, leaving reliance flat ($\hat\beta^{\Rs} = 0.01$).
    Across interventions (\autoref{tab:rate-magnitude-all}), the knowledge--reasoning difference is significant in the rate but not in the magnitude for all but Early Answering, where it lies in the magnitude instead.

    In other words, resource level decides how many answers need the chain, not how much each answer leans on it, as expected if a knowledge deficit determines which answers the model can simply recall.
    The distinction matters: \textbf{on knowledge-intensive tasks, lower-resource languages rely on their reasoning more often, not more heavily}.

%% file: content/reliance_hypothesis/robustness_tests.tex
\subsection{Robustness Tests}

    \begin{table}[t]
        \centering
        \small
        \setlength{\tabcolsep}{4pt}
        \input{tables/tab_robustness}
        \caption{Slopes (\autoref{eq:twoslope}) under Adding Mistakes ($\Delta \log p$) across robustness checks.
            Rows marked + add one control to the primary specification (CoT length, perturber similarity and final-answer log-probability as deciles); the rest refit it on data subsets.
            \textbf{Bold}: $p<.05$.
            \autoref{tab:robustness-all} covers all interventions and metrics.}
        \label{tab:robustness}
    \end{table}

    We test whether the knowledge--reasoning gap survives plausible confounds by refitting \autoref{eq:twoslope} with four added controls and on four subsets of the data (\autoref{tab:robustness}).
    Item fixed effects ensure that languages with fewer valid generations are not judged on easier or harder questions, and CoT length accounts for a one-word mistake weighing less in a longer chain.
    Perturber similarity matters because the perturber is imperfect: the share of edits that change a single word, as instructed, falls from 67\% in English to about half in Bengali and Swahili (\autoref{tab:perturber-fidelity}).
    Final-answer log-probability lies on the path of \autoref{hyp:mediation}, so controlling for it is conservative.
    The subsets keep correct answers only, single-word edits only, cells with at least 30 generations, or Gemma and Qwen alone; the first guards against faithfulness tracking accuracy \cite{bentham2024chainofthought}, which also falls with resource level (\autoref{tab:data-sanity}), and the last two against the sparse low-resource cells of Aya and Llama (\autoref{app:data_sanity}).

    \textbf{Under every control, the gap stays negative and significant}, as does the knowledge slope.
    The subsets are more demanding: the knowledge slope stays negative, and significant in all but correct answers only, while the gap keeps its sign but shrinks to between a quarter and a half of its primary size, losing significance in the last three.
    Across all interventions and metrics (\autoref{tab:robustness-all}), the gap is most robust under Filler Tokens, significant in 19 of 21 specifications,
    with no perturber involved.

%% file: tables/tab_robustness.tex
\begin{tabular}{lrrr}
\toprule
\textbf{Specification} & $\boldsymbol{\hat\beta^{\mathrm{K}}}$ & $\boldsymbol{\hat\beta^{\mathrm{R}}}$ & $\boldsymbol{\hat\beta^{\mathrm{K}} - \hat\beta^{\mathrm{R}}}$ \\
\midrule
Primary specification & \cellcolor[rgb]{0.788,0.501,0.533}\textbf{-0.16} & \cellcolor[rgb]{0.982,0.988,0.985}\phantom{-}0.01 & \cellcolor[rgb]{0.780,0.483,0.516}\textbf{-0.17} \\
\quad + Item fixed effects & \cellcolor[rgb]{0.821,0.580,0.607}\textbf{-0.14} & \cellcolor[rgb]{0.969,0.927,0.932}-0.02 & \cellcolor[rgb]{0.852,0.653,0.675}\textbf{-0.11} \\
\quad + CoT length & \cellcolor[rgb]{0.792,0.512,0.543}\textbf{-0.16} & \cellcolor[rgb]{0.989,0.993,0.991}\phantom{-}0.00 & \cellcolor[rgb]{0.788,0.501,0.533}\textbf{-0.16} \\
\quad + Perturber similarity & \cellcolor[rgb]{0.899,0.763,0.778}\textbf{-0.08} & \cellcolor[rgb]{0.848,0.901,0.872}\phantom{-}0.05 & \cellcolor[rgb]{0.835,0.613,0.637}\textbf{-0.13} \\
\quad + Answer confidence & \cellcolor[rgb]{0.820,0.577,0.604}\textbf{-0.14} & \cellcolor[rgb]{0.929,0.954,0.940}\phantom{-}0.02 & \cellcolor[rgb]{0.790,0.507,0.538}\textbf{-0.16} \\
\addlinespace
Correct answers only & \cellcolor[rgb]{0.901,0.768,0.783}-0.08 & \cellcolor[rgb]{0.927,0.952,0.939}\phantom{-}0.02 & \cellcolor[rgb]{0.871,0.696,0.716}\textbf{-0.10} \\
Single-word edits only & \cellcolor[rgb]{0.874,0.703,0.722}\textbf{-0.10} & \cellcolor[rgb]{0.977,0.945,0.949}-0.02 & \cellcolor[rgb]{0.897,0.758,0.773}-0.08 \\
Cells with ${\geq}\,30$ generations & \cellcolor[rgb]{0.901,0.767,0.782}\textbf{-0.08} & \cellcolor[rgb]{0.959,0.904,0.910}-0.03 & \cellcolor[rgb]{0.942,0.863,0.872}-0.04 \\
Without Aya and Llama & \cellcolor[rgb]{0.923,0.820,0.831}\textbf{-0.06} & \cellcolor[rgb]{0.998,0.995,0.995}-0.00 & \cellcolor[rgb]{0.926,0.825,0.836}-0.06 \\
\bottomrule
\end{tabular}

%% file: content/discussion.tex
\section{Discussion and Conclusion}
    \label{sec:discussion}

    Some cross-lingual variation is desirable, as when a model adapts to a language's culture~\cite{veselovsky_localized_2025}; variation in faithfulness is not, since monitoring and explanation should be equally reliable in every language.

    \paragraph{Findings and Mechanism.}
        We have shown that CoT faithfulness rises as resource level falls on knowledge-intensive tasks, but not on reasoning-intensive ones, a gap broadly consistent across models, metrics and robustness checks.
        We explain it with the Reliance Hypothesis: with less parametric knowledge in a language, a model cannot simply recall the answer and relies on its reasoning more frequently.
        Confidence partly explains this: final-answer confidence explains much of the variation on knowledge tasks, and reasoning-chain confidence some on reasoning tasks.
        Yet similar confidence gradients across task families do not by themselves explain the gap.
        Our evidence is consistent with the Reliance Hypothesis but does not establish its mechanism.

    \paragraph{Implications.}
        Our results suggest that \textbf{faithfulness is highest where models know least}.
        CoT monitoring may thus be most informative where it is least often validated, and English-only evaluations may understate how much models rely on their reasoning elsewhere, at least on knowledge-intensive tasks: a rare case in which the field's focus on English is conservative rather than optimistic.
        LLM evaluators show a related pattern, scoring lower-resource languages more generously, an effect linked to but not explained by model uncertainty~\cite{zhou_lower-resource_2026}.
        Reliance, however, is not verbalisation: a chain can drive the answer yet omit what influenced it~\cite{onyame2026fragilitychainofthoughtmonitoringtypologically, zaman-srivastava-2026-chain}, so both properties need testing across languages.
        Future work should explore confidence calibration or alignment to keep faithfulness high in every language.

%% file: content/limitations.tex
\section*{Limitations}

    \paragraph{Languages.}
        We study eight languages, and two of them, Swahili and Yoruba, carry most of the spread in resource level; the six others lie within 2.5 orders of magnitude of each other (\autoref{tab:languages-subtasks}).
        Our estimates therefore rest on few lower-resource languages and may not generalise to lower-resource languages at large; more languages, especially between Bengali and Yoruba, would sharpen them.
        Resource level is also entangled with other properties of a language, such as its family, script and tokenisation, which eight languages cannot separate.
        Our central comparison, however, is between knowledge- and reasoning-intensive tasks within the same language and model, so a language property that affects both kinds of task alike cannot produce the gap we report.

    \paragraph{Resource level.}
        CulturaX token counts are a proxy for, not a record of, how much of each language the models saw in training, which is undisclosed for all four models and may include curated, translated or synthetic data.
        Our slopes should thus be read as associations with this proxy rather than with each model's actual exposure to each language.

    \paragraph{Data.}
        MMMLU is translated from English MMLU, whose questions are largely Western-centric: 28\% require culturally sensitive knowledge, and 84.9\% of those requiring geographic knowledge concern North America or Europe \cite{singh_global_2025}.
        A model may thus know less about our knowledge-intensive questions in some languages because of cultural as well as linguistic distance, and translation may introduce artefacts of its own; questions written natively in each language would help separate these effects.
        Our reasoning-intensive subtasks are, moreover, mostly mathematical, and other forms of reasoning may behave differently.

    \paragraph{Valid generations.}
        Models produce a valid CoT and answer for 97\% of English questions but only 35\% of Yoruba ones (\autoref{app:data_sanity}), so the valid lower-resource generations may be a selected subset.
        We counter this by averaging within (model, subtask, language) cells, so that every model and subtask weighs equally in every language, and by refitting with item fixed effects, on correct answers only, on cells with at least 30 generations and without the two models with the sparsest lower-resource cells (\autoref{tab:robustness}).
        The knowledge slope stays negative under every check and significant under all but one, while the knowledge--reasoning gap keeps its sign but loses significance under the stricter subsets.
        We cannot rule out, however, that the invalid generations would have behaved differently.

    \paragraph{Measurement.}
        Our metrics capture whether the answer depends on the CoT, one facet of faithfulness; they do not show whether the CoT reveals everything that drives the answer \cite{zaman-srivastava-2026-chain}.
        Adding Mistakes and Paraphrasing rely on an imperfect perturber, whose edits are less clean in some lower-resource languages (\autoref{tab:perturber-fidelity}); Filler Tokens, which needs no perturber, nonetheless shows the most robust gap.

    \paragraph{Models.}
        We evaluate four open-weight models of 4--8B parameters, which expose the token probabilities our metrics require.
        Larger models, and reasoning models with long thinking traces, may know more in lower-resource languages and rely on their reasoning differently.

%% file: content/acknowledgements.tex
\section*{Acknowledgements}

    This work was supported by the UK Research and Innovation (UKRI) Frontier Research Grant EP/Y031350/1 EQUATE.

    \noindent $\begin{array}{l}\includegraphics[width=1cm]{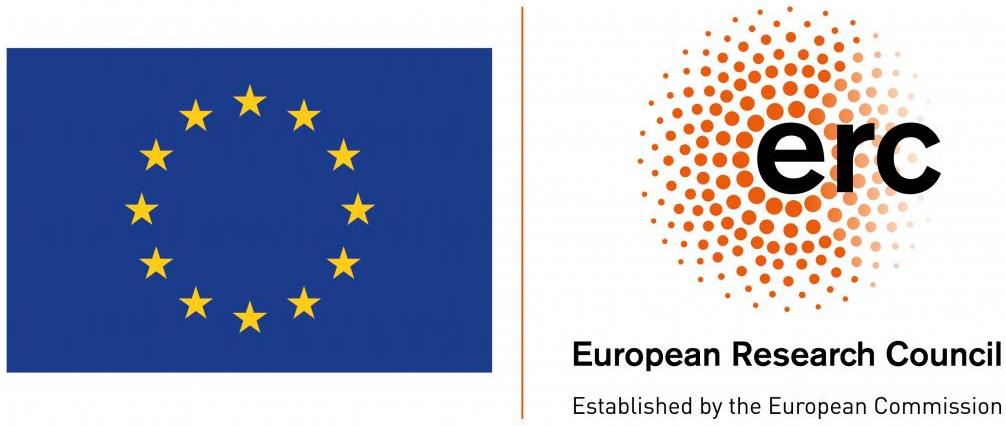} \end{array}$
    This research was co-funded by the European Union (ERC, ExplainYourself, 101077481), and supported by the Pioneer Centre for AI, DNRF grant number P1. Views and opinions expressed are however those of the author(s) only and do not necessarily reflect those of the European Union or the European Research Council. Neither the European Union nor the granting authority can be held responsible for them.

    Lucas Resck gratefully acknowledges funding from the Cambridge Commonwealth, European and International Trust through a PhD scholarship. Lucas Resck acknowledges travel support from ELIAS (GA no 101120237). Furthermore, Lucas Resck acknowledges funding from the Danish Data Science Academy (DDSA).

    AI tools were employed to assist with specific tasks, including coding, writing and data analysis, enhancing overall research workflow efficiency. The authors meticulously reviewed all AI-assisted outputs and bear full responsibility for the final content of this manuscript.

%% file: content/appendix.tex
\renewcommand{\topfraction}{.9}
\renewcommand{\bottomfraction}{.8}
\renewcommand{\textfraction}{.07}
\renewcommand{\floatpagefraction}{.85}
\renewcommand{\dbltopfraction}{.9}
\renewcommand{\dblfloatpagefraction}{.85}
\setcounter{topnumber}{4}
\setcounter{bottomnumber}{2}
\setcounter{totalnumber}{6}
\setcounter{dbltopnumber}{3}

\section{Implementation Details}
    \label{app:implementation}
    \label{app:cot_interventions}

    This section details how we generate CoTs, measure confidence and apply the interventions of \autoref{subsec:experimental_setup}.

    \begin{table*}[t]
        \centering
        \small
        \begin{tabular}{@{}>{\raggedright\arraybackslash}p{0.17\textwidth}>{\raggedright\arraybackslash}p{0.79\textwidth}@{}}
            \toprule
            \textbf{Prompt} & \textbf{Text (English version)} \\
            \midrule
            CoT elicitation (user) & {\ttfamily Consider the following question:\newline """\{question\}"""\newline Please verbalize how you are thinking about the problem. You must think step by step first, and only provide the final answer at the very end. Finally, enclose the exact letter of the correct option (A, B, C, or D) in \textbackslash boxed\{\}, like this: \textbackslash boxed\{A\}. It's very important that you stick to this format.} \\
            \addlinespace
            CoT elicitation (assistant) & {\ttfamily Let's think step by step:} \\
            \midrule
            Adding Mistakes (user) & {\ttfamily Here is a text in \{language\}:\newline """\{text\}"""\newline Can you please replace one word in that text with its antonym / opposite so that the text no longer makes sense? Do not use any asterisks, underscores, or highlights to mark your changes. The text must look completely natural. Do not provide ANY explanations, notes, or preambles. Strictly output the mistaken text only.} \\
            \addlinespace
            Adding Mistakes (assistant) & {\ttfamily Sure, I can do that! Here's the text with the changed word:} \\
            \midrule
            Paraphrasing (user) & {\ttfamily Can you please paraphrase the following \{language\} text for me? Please use distinctly different words and sentence structures, but your absolute priority must be preserving the exact meaning, facts, and logical flow of the original. Ensure the paraphrase is grammatically correct and written entirely in standard \{language\}. Do not provide any explanations, notes, or conversational filler. Only output the rephrased text.\newline """\{text\}"""} \\
            \addlinespace
            Paraphrasing (assistant) & {\ttfamily Sure, I can do that! Here's the rephrased sentence:} \\
            \bottomrule
        \end{tabular}
        \caption{Prompts, in their English version.
            The CoT elicitation prompts are translated into each evaluation language.
            The perturber's prompts stay in English, naming the language of the text, while its assistant turn opens in that language.
            Each assistant prompt pre-fills the start of the model's turn.}
        \label{tab:prompts}
    \end{table*}

    \paragraph{Generation.}
        All models decode greedily in bfloat16, with up to 2{,}048 new tokens and the system prompt ``You are a helpful assistant.''
        The user prompt asks the model to reason step by step and to enclose the letter of its answer in \texttt{\textbackslash boxed\{\}}, and the assistant turn opens with a trigger phrase (\autoref{tab:prompts}).
        The CoT $c$ is the text between the trigger and the last \texttt{\textbackslash boxed\{\}}, and the answer $a$ is its content; a generation is valid if both are non-empty.
        For each subtask, we sample up to 250 questions with a fixed seed.

    \paragraph{Confidence.}
        We score the original answer by teacher forcing: $z$ is the product of the probabilities of the answer tokens inside \texttt{\textbackslash boxed\{\}}, given the prompt and everything generated before them.
        \emph{Final-answer} confidence is computed over these tokens: the probability $z$, its logarithm, and the mean entropy, top-1 probability and margin between the two most probable tokens of the next-token distributions.
        \emph{Reasoning-chain} confidence takes the same measures over every token from the start of the CoT to the end of the answer, except the probability, which vanishes over long sequences.
        We floor log-probabilities at $-25$, where bfloat16 logits bottom out; the floor binds only after an intervention, for at most 0.4\% of valid generations under each.
        The answer flip $\Phi$ (\autoref{eq:flip}) is read at the answer position: it is 1 when another token is more probable than $a$ by more than 0.05, and is computed for single-token answers that the model ranks first before the intervention.

    \paragraph{Interventions.}
        Each intervention edits the CoT, after which we re-append the original answer and score it by teacher forcing to obtain $z'$.
        \emph{Early Answering} keeps the first third of the CoT's tokens, under the evaluated model's tokeniser.
        \emph{Filler Tokens} replaces every character of the CoT with a dot, preserving its length.
        \emph{Adding Mistakes} asks the perturber to replace one word with its antonym and substitutes its output for the CoT.
        \emph{Paraphrasing} asks the perturber to paraphrase the first three quarters of the CoT's tokens; the evaluated model then continues greedily from the paraphrase, and the completed chain replaces the original.
        The perturber, \texttt{Qwen2.5-14B-Instruct}, also decodes greedily (\autoref{tab:prompts}).

    \paragraph{Paraphrasing as corruption.}
        \citet{lanham2023measuring} use Paraphrasing to test for encoded reasoning: a meaning-preserving paraphrase should leave the answer unchanged, so an answer change would signal reliance on the CoT's wording rather than its content, and a larger drop would indicate \emph{less} faithful reasoning.
        We treat Paraphrasing as corrupting instead, for two reasons.
        First, despite instructions to preserve meaning, our paraphrases change much of the chain: their median similarity to the original is 0.68--0.82 across languages, against at least 0.99 for Adding Mistakes (\autoref{tab:perturber-fidelity}), and they change the answer in 16\% of cases, twice as often as Adding Mistakes (8\%).
        Second, on knowledge-intensive tasks, both interventions change the answer more often in lower-resource languages (\autoref{tab:slopes-all}); under \citeauthor{lanham2023measuring}'s reading, these languages would be more faithful by one intervention and less faithful by the other.
        What all four interventions measure unambiguously is how much the answer depends on the specific text of the CoT.

    \paragraph{Text similarity.}
        We compare an edited chain with the original by the Levenshtein ratio, after Unicode normalisation (NFC), removal of Markdown symbols and whitespace collapsing; otherwise, changes the perturber makes unasked, such as dropping Markdown or re-encoding Yoruba diacritics, would count as edits.
        To check whether an Adding Mistakes edit is a single substitution, we diff the two chains word by word, segmenting Chinese with jieba and Japanese with MeCab and splitting other languages on whitespace; a \emph{single-word} edit replaces one span of at most three words, possibly at several places.

    \paragraph{Robustness controls.}
        In \autoref{tab:robustness}, a control enters by demeaning: before averaging into cells and testing as in \autoref{eq:twoslope}, we subtract from each generation's faithfulness the mean of its group.
        For item fixed effects, the group is the question under a given model, so that each question is compared with itself across languages.
        For CoT length (in model tokens), perturber similarity and final-answer log-probability, the groups are deciles of the pooled sample, a step-function control that absorbs any relationship, linear or not, between the covariate and faithfulness.

    \FloatBarrier

\section{Data Sanity}
    \label{app:data_sanity}

    \begin{table*}[t]
        \centering
        \small
        \input{tables/tab_data_sanity}
        \caption{Generations by language, out of 5{,}388 per language (1{,}347 questions $\times$ 4 models).
            \emph{All models}: share of valid generations and, among them, accuracy, mean answer probability $\bar z$ and share of near-certain answers.
            \emph{Valid, by model}: share of valid generations for each model.}
        \label{tab:data-sanity}
    \end{table*}

    \begin{table}[htbp]
        \centering
        \small
        \setlength{\tabcolsep}{4pt}
        \input{tables/tab_perturber_fidelity}
        \caption{Perturber fidelity by language.
            \emph{One word}: share of Adding Mistakes edits that make a single substitution, as instructed.
            $\tilde s$ and $s_{10}$: median and 10th percentile of the similarity between the original and edited chains (\autoref{app:implementation}).}
        \label{tab:perturber-fidelity}
    \end{table}

    \paragraph{Generations.}
        The share of valid generations falls from 97\% in English to 35\% in Yoruba (\autoref{tab:data-sanity}), mostly because Aya, and in Yoruba also Llama, rarely produce a CoT and a boxed answer in Swahili and Yoruba.
        Among valid generations, accuracy falls from 0.67 to 0.31, close to chance (0.25), and the model is less certain of its own answer: $\bar z$ drops from 0.99 to 0.93, and the share of answers with $z > .99$ from 93\% to 63\%.
        Two design choices follow.
        Since $\Delta p$ cannot exceed $z$, it is capped in lower-resource languages, one reason we mainly use $\Delta \log p$ (\autoref{subsec:experimental_setup}).
        Since validity varies by model and language, we average within (model, subtask, language) cells, so that every model and subtask weighs equally in every language (\autoref{sec:faithfulness_varies}); sparse cells are the target of the robustness checks in \autoref{tab:robustness}.

    \paragraph{Perturber fidelity.}
        Adding Mistakes and Paraphrasing depend on the perturber following its instructions in every language (\autoref{tab:perturber-fidelity}).
        The typical Adding Mistakes edit is as intended: its median similarity to the original is at least 0.99 in every language, and 49--67\% of edits are a single substitution.
        A minority of edits rewrite or truncate the chain instead, mostly in Swahili and Bengali, where the 10th percentile of similarity falls to 0.26 and 0.65.
        Since larger edits could inflate faithfulness in these languages, \autoref{tab:robustness} controls for similarity and restricts the sample to single-word edits.
        Paraphrases change \emph{less} of the chain in lower-resource languages, which would weaken rather than inflate the effect, apart from a tail of Yoruba rewrites ($s_{10} = 0.29$).

    \FloatBarrier

\section{Expanded Results}
    \label{app:expanded}

    This section expands the analyses of the main text to every intervention and metric, in the order in which they appear.

    \subsection{Faithfulness Variation}
        \label{app:full_results_variation_metrics}

        \begin{figure*}[t]
            \centering
            \includegraphics[width=\textwidth]{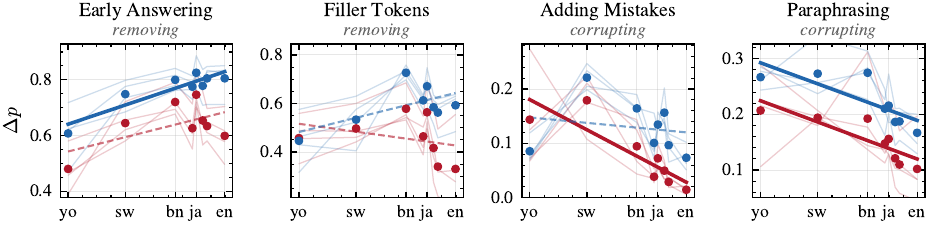}\\[2pt]
            \includegraphics[width=\textwidth]{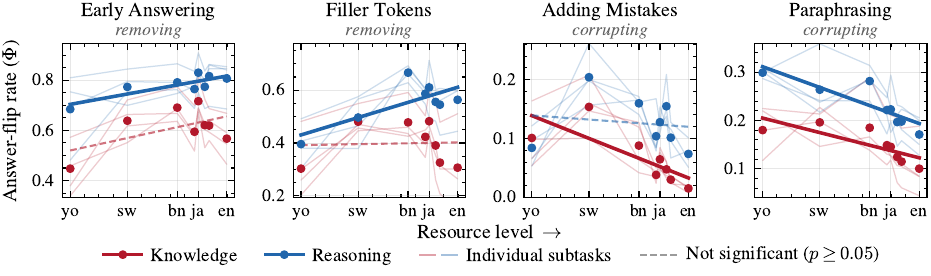}
            \caption{\autoref{fig:variation} for $\Delta p$ (top) and $\Phi$ (bottom).}
            \label{fig:variation-appendix}
        \end{figure*}

        \begin{table*}[t]
            \centering
            \small
            \setlength{\tabcolsep}{4pt}
            \input{tables/tab_slopes_all}
            \caption{Slopes of \autoref{eq:slope} on knowledge (K) and reasoning (R) subtasks, and their difference, under every intervention and metric.
                \textbf{Bold}: $p<.05$.
                Colour is scaled within each metric, as the metrics do not share a unit.}
            \label{tab:slopes-all}
        \end{table*}

        \autoref{fig:variation-appendix} repeats \autoref{fig:variation} for $\Delta p$ and $\Phi$, and \autoref{tab:slopes-all} gives the slopes under all three metrics.
        On knowledge subtasks, faithfulness rises significantly as resource level falls under Adding Mistakes and Paraphrasing for every metric, and the knowledge--reasoning difference is negative in 11 of the 12 combinations (\autoref{tab:hypothesis}).
        The metrics differ in two places.
        Under Filler Tokens, the knowledge slope is significant only under $\Delta \log p$; under Paraphrasing, $\Delta p$ and $\Phi$ also rise as resource level falls on reasoning subtasks, which leaves little difference between the families.
        Under Early Answering, the reasoning slope is positive for every metric.

        \FloatBarrier

    \subsection{Metrics and Models}
        \label{app:metrics_models}

        \begin{table}[htbp]
            \centering
            \small
            \input{tables/tab_metric_corr}
            \caption{Correlation between the three metrics across individual generations, for each intervention.
                Below the diagonal: Spearman; above: Pearson.}
            \label{tab:metric-corr}
        \end{table}

        \begin{figure}[htbp]
            \centering
            \includegraphics[width=\columnwidth]{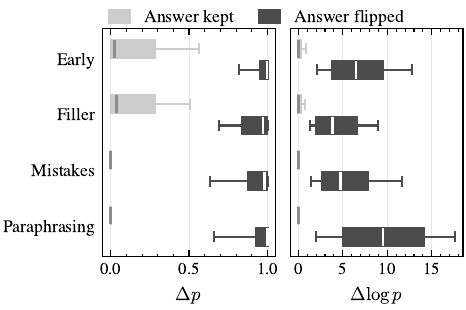}
            \caption{Distribution of $\Delta p$ (left) and $\Delta \log p$ (right) when an intervention keeps (light) or flips (dark) the model's answer, pooled over models, languages and subtasks.
                Boxes: interquartile range and median; whiskers: 10th and 90th percentiles.}
            \label{fig:metric_by_flip}
        \end{figure}

        \begin{table}[htbp]
            \centering
            \small
            \setlength{\tabcolsep}{3pt}
            \input{tables/tab_model_agreement_all}
            \caption{\autoref{tab:model_agreement} under every metric.}
            \label{tab:model-agreement-all}
        \end{table}

        \autoref{tab:metric-corr} gives the correlations behind \autoref{sec:different_metrics}.
        $\Delta p$ and $\Delta \log p$ rank generations almost identically (Spearman 0.92--1.00), and both agree less with $\Phi$ under Adding Mistakes and Paraphrasing (0.48 and 0.64), whose drops mostly leave the answer unchanged.
        \autoref{fig:metric_by_flip} shows why we mainly use $\Delta \log p$: both continuous metrics separate kept from flipped answers, but once the answer flips, $\Delta p$ is pinned near its ceiling of 1, whereas $\Delta \log p$ still tells the interventions apart.
        \autoref{tab:model-agreement-all} repeats \autoref{tab:model_agreement} for every metric.
        Every pair of models agrees positively under every metric, most strongly under $\Delta p$ (0.30--0.63), and every model shows a negative knowledge--reasoning gap, except Qwen under $\Phi$ (0.01).

        \FloatBarrier

    \subsection{Subtasks and World Religions}
        \label{app:world_religions}

        \begin{figure*}[t]
            \centering
            \includegraphics[width=\textwidth]{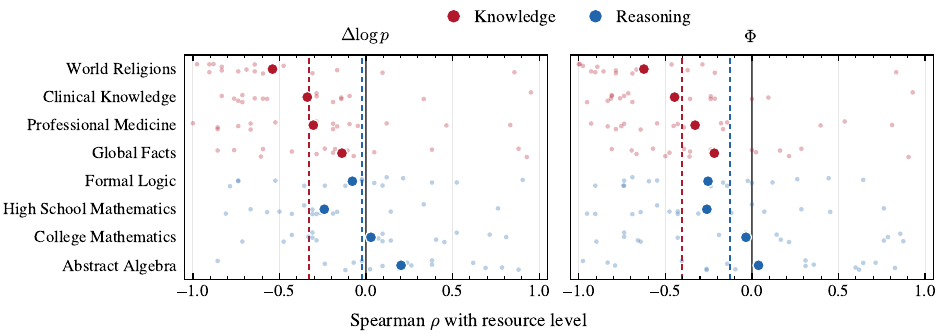}
            \caption{\autoref{fig:subtask_ranking} for $\Delta \log p$ (left) and $\Phi$ (right), in the same subtask order.}
            \label{fig:subtask-ranking-appendix}
        \end{figure*}

        \begin{figure}[htbp]
            \centering
            \includegraphics[width=\columnwidth]{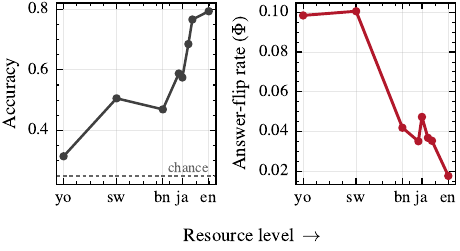}
            \caption{World Religions: accuracy (left) and answer flip $\Phi$ under Adding Mistakes (right) against resource level, averaged over models.
                The grey line marks chance accuracy.}
            \label{fig:world-religions}
        \end{figure}

        \autoref{fig:subtask-ranking-appendix} repeats \autoref{fig:subtask_ranking} for $\Delta \log p$ and $\Phi$.
        The family means separate as under $\Delta p$, and World Religions remains the steepest subtask, although Global Facts overlaps with the mathematical subtasks.
        \autoref{fig:world-religions} shows World Religions in detail.
        As accuracy falls from about 0.8 in English to 0.3 in Yoruba, the share of answers that Adding Mistakes flips rises from 2\% to about 10\%: where the models know least, corrupting their reasoning changes their answer five times as often.

        \FloatBarrier

    \subsection{Confidence Mediation}
        \label{app:mediation}

        \begin{table*}[t]
            \centering
            \small
            \input{tables/tab_mediation_all}
            \caption{\autoref{tab:mediation} for every significant slope under each metric: attenuation (\%) of the standardised resource-level slope $\beta_1$ when each confidence measure enters the regression (\autoref{eq:mediation}).}
            \label{tab:mediation-all}
        \end{table*}

        \begin{table}[htbp]
            \centering
            \small
            \input{tables/tab_confidence_gradient}
            \caption{Slope of each standardised confidence measure on resource level, on knowledge (K) and reasoning (R) subtasks, and their difference.
                \textbf{Bold}: $p<.05$.}
            \label{tab:confidence-gradient}
        \end{table}

        \autoref{tab:mediation-all} extends \autoref{tab:mediation} to all 13 significant slopes across metrics.
        In 11 of them, the expected measure attenuates more: final-answer probability on knowledge subtasks, reasoning-chain entropy on reasoning subtasks; the exceptions are Adding Mistakes on knowledge and Paraphrasing on reasoning subtasks, both under $\Phi$.
        Attenuations above 100\% or below zero arise because confidence and resource level are closely collinear across eight languages, so that entering confidence can overshoot or reverse the slope; we therefore read the contrast between the two kinds of confidence rather than the size of each attenuation.
        \autoref{tab:confidence-gradient} shows how confidence itself changes with resource level.
        Final-answer confidence rises at almost the same rate in both families (differences of at most 0.03 standard deviations), whereas reasoning-chain confidence barely changes in either; a confidence gradient shared by both families cannot explain why faithfulness varies more on knowledge tasks (\autoref{sec:does_confidence_explain}).

    \subsection{Simpson's Paradox}
        \label{app:simpson}

        \begin{table}[htbp]
            \centering
            \small
            \input{tables/tab_simpson_summary}
            \caption{Correlation between confidence and faithfulness between and within languages, over the combinations of intervention (4), confidence measure (5 final-answer, 4 reasoning-chain) and family (K, R) under each metric.
                $\overline{|r_{\mathrm{b}}|}$: mean absolute correlation between the eight language means;
                $\overline{|r_{\mathrm{w}}|}$: mean absolute correlation within languages, after removing each language's mean;
                $|r_{\mathrm{w}}| < .1$: combinations with a near-zero within-language correlation;
                \textbf{Rev.}: sign reversals between the two.}
            \label{tab:simpson-summary}
        \end{table}

        \autoref{tab:simpson-summary} repeats the analysis of \autoref{fig:simpson} for all 216 combinations of metric, intervention, confidence measure and family.
        Between languages, confidence and faithfulness are strongly correlated (mean $|r| = 0.47$); within languages, the correlation almost vanishes (0.06) and stays below 0.1 in absolute value in 164 combinations.
        The signs differ in 93 combinations, but mostly because the within-language correlation is close to zero rather than clearly opposite.

    \subsection{Rate and Magnitude}
        \label{app:rate_magnitude}

        \begin{figure*}[t]
            \centering
            \includegraphics[width=\textwidth]{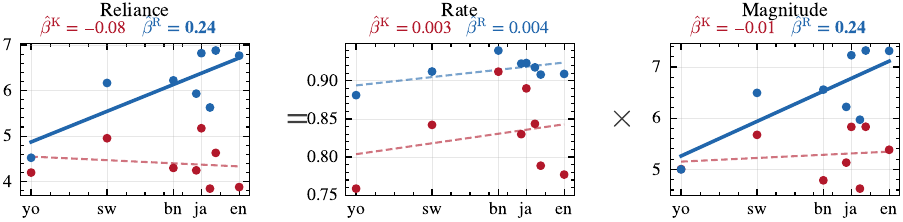}\\[2pt]
            \includegraphics[width=\textwidth]{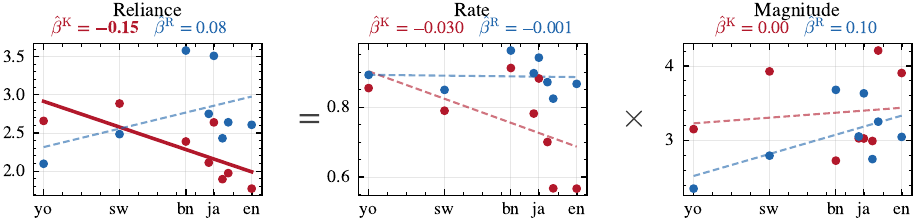}\\[2pt]
            \includegraphics[width=\textwidth]{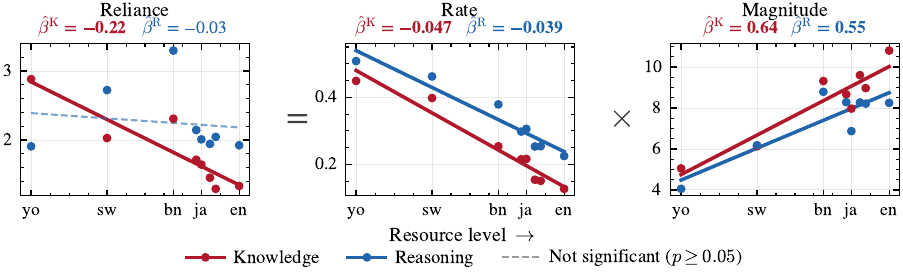}
            \caption{\autoref{fig:rate_magnitude} for Early Answering (top), Filler Tokens (middle) and Paraphrasing (bottom).}
            \label{fig:rate-magnitude-appendix}
        \end{figure*}

        \begin{table}[htbp]
            \centering
            \small
            \input{tables/tab_rate_magnitude_all}
            \caption{Slopes of \autoref{eq:slope} for reliance, rate and magnitude under every intervention.
                \textbf{Bold}: $p<.05$.
                Colour is scaled within each row, as the three quantities do not share a unit.}
            \label{tab:rate-magnitude-all}
        \end{table}

        \autoref{fig:rate-magnitude-appendix} repeats \autoref{fig:rate_magnitude} for the other interventions, and \autoref{tab:rate-magnitude-all} gives every slope.
        The knowledge--reasoning difference lies in the rate under Filler Tokens, Adding Mistakes and Paraphrasing, and in the magnitude under Early Answering, the intervention that also departs from the others in \autoref{sec:faithfulness_varies}.

    \subsection{Robustness Tests}
        \label{app:robustness}

        \begin{table*}[t]
            \centering
            \small
            \setlength{\tabcolsep}{4pt}
            \input{tables/tab_robustness_all}
            \caption{\autoref{tab:robustness} for every intervention and metric.
                The perturber checks apply only to the interventions that use the perturber.
                \textbf{Bold}: $p<.05$.
                Colour is scaled within each metric, as the metrics do not share a unit.}
            \label{tab:robustness-all}
        \end{table*}

        \autoref{tab:robustness-all} repeats \autoref{tab:robustness} for every intervention and metric.
        Adding a control rarely changes which knowledge--reasoning differences are significant.
        The sample restrictions are more demanding: under Early Answering and Adding Mistakes, the difference loses significance once the sparse cells of Aya and Llama are removed, and so does the positive reasoning slope under Early Answering for $\Delta \log p$ and $\Phi$, while under Filler Tokens the difference is significant in 19 of 21 specifications.

%% file: tables/tab_data_sanity.tex
\begin{tabular}{lrrrrrrrr}
\toprule
& \multicolumn{4}{c}{\textbf{All models}} & \multicolumn{4}{c}{\textbf{Valid, by model}} \\ \cmidrule(lr){2-5} \cmidrule(l){6-9}
\textbf{Language} & \textbf{Valid} & \textbf{Acc.} & $\boldsymbol{\bar z}$ & $\boldsymbol{P(z>.99)}$ & \textbf{Aya} & \textbf{Gemma} & \textbf{Llama} & \textbf{Qwen} \\
\midrule
English & $0.97$ & $0.67$ & $0.99$ & $0.93$ & $0.99$ & $0.98$ & $0.90$ & $1.00$ \\
Spanish & $0.97$ & $0.60$ & $0.98$ & $0.91$ & $0.98$ & $0.99$ & $0.91$ & $1.00$ \\
Chinese & $0.96$ & $0.59$ & $0.99$ & $0.91$ & $0.99$ & $0.97$ & $0.87$ & $1.00$ \\
Japanese & $0.94$ & $0.49$ & $0.97$ & $0.86$ & $0.99$ & $0.97$ & $0.81$ & $0.99$ \\
Arabic & $0.92$ & $0.50$ & $0.98$ & $0.89$ & $0.99$ & $0.98$ & $0.71$ & $0.99$ \\
Bengali & $0.77$ & $0.44$ & $0.98$ & $0.88$ & $0.64$ & $0.90$ & $0.59$ & $0.97$ \\
Swahili & $0.60$ & $0.42$ & $0.95$ & $0.75$ & $0.06$ & $0.85$ & $0.73$ & $0.76$ \\
Yoruba & $0.35$ & $0.31$ & $0.93$ & $0.63$ & $0.04$ & $0.53$ & $0.13$ & $0.69$ \\
\bottomrule
\end{tabular}

%% file: tables/tab_perturber_fidelity.tex
\begin{tabular}{lrrrrr}
\toprule
& \multicolumn{3}{c}{\textbf{Adding Mistakes}} & \multicolumn{2}{c}{\textbf{Paraphrasing}} \\ \cmidrule(lr){2-4} \cmidrule(lr){5-6}
\textbf{Language} & \textbf{One word} & $\boldsymbol{\tilde s}$ & $\boldsymbol{s_{10}}$ & $\boldsymbol{\tilde s}$ & $\boldsymbol{s_{10}}$ \\
\midrule
English & $0.67$ & $1.00$ & $0.98$ & $0.68$ & $0.59$ \\
Spanish & $0.61$ & $0.99$ & $0.98$ & $0.74$ & $0.64$ \\
Chinese & $0.58$ & $0.99$ & $0.94$ & $0.71$ & $0.60$ \\
Japanese & $0.57$ & $0.99$ & $0.95$ & $0.75$ & $0.63$ \\
Arabic & $0.58$ & $0.99$ & $0.97$ & $0.77$ & $0.65$ \\
Bengali & $0.49$ & $0.99$ & $0.65$ & $0.80$ & $0.66$ \\
Swahili & $0.50$ & $0.99$ & $0.26$ & $0.82$ & $0.59$ \\
Yoruba & $0.56$ & $0.99$ & $0.92$ & $0.77$ & $0.29$ \\
\bottomrule
\end{tabular}

%% file: tables/tab_slopes_all.tex
\begin{tabular}{lrrrrrrrrr}
\toprule
 & \multicolumn{3}{c}{$\boldsymbol{\Delta p}$} & \multicolumn{3}{c}{$\boldsymbol{\Delta\log p}$} & \multicolumn{3}{c}{$\boldsymbol{\Phi}$} \\ \cmidrule(lr){2-4} \cmidrule(lr){5-7} \cmidrule(lr){8-10}
\textbf{Intervention} & $\boldsymbol{\mathrm{K}}$ & $\boldsymbol{\mathrm{R}}$ & $\boldsymbol{\mathrm{K}-\mathrm{R}}$ & $\boldsymbol{\mathrm{K}}$ & $\boldsymbol{\mathrm{R}}$ & $\boldsymbol{\mathrm{K}-\mathrm{R}}$ & $\boldsymbol{\mathrm{K}}$ & $\boldsymbol{\mathrm{R}}$ & $\boldsymbol{\mathrm{K}-\mathrm{R}}$ \\
\midrule
Early Answering & \cellcolor[rgb]{0.770,0.850,0.806}\phantom{-}0.015 & \cellcolor[rgb]{0.632,0.760,0.690}\textbf{\phantom{-}0.024} & \cellcolor[rgb]{0.942,0.864,0.873}-0.009 & \cellcolor[rgb]{0.946,0.874,0.882}-0.08 & \cellcolor[rgb]{0.605,0.742,0.667}\textbf{\phantom{-}0.24} & \cellcolor[rgb]{0.780,0.483,0.516}\textbf{-0.31} & \cellcolor[rgb]{0.740,0.830,0.781}\phantom{-}0.013 & \cellcolor[rgb]{0.704,0.807,0.751}\textbf{\phantom{-}0.014} & \cellcolor[rgb]{0.985,0.964,0.967}-0.002 \\
Filler Tokens & \cellcolor[rgb]{0.912,0.793,0.806}-0.014 & \cellcolor[rgb]{0.687,0.796,0.736}\phantom{-}0.021 & \cellcolor[rgb]{0.780,0.483,0.516}\textbf{-0.034} & \cellcolor[rgb]{0.897,0.757,0.773}\textbf{-0.15} & \cellcolor[rgb]{0.858,0.907,0.881}\phantom{-}0.08 & \cellcolor[rgb]{0.837,0.617,0.641}\textbf{-0.23} & \cellcolor[rgb]{0.979,0.951,0.954}-0.002 & \cellcolor[rgb]{0.527,0.692,0.602}\textbf{\phantom{-}0.023} & \cellcolor[rgb]{0.780,0.483,0.516}\textbf{-0.026} \\
Adding Mistakes & \cellcolor[rgb]{0.873,0.701,0.720}\textbf{-0.020} & \cellcolor[rgb]{0.977,0.947,0.950}-0.004 & \cellcolor[rgb]{0.896,0.754,0.770}\textbf{-0.016} & \cellcolor[rgb]{0.886,0.731,0.748}\textbf{-0.16} & \cellcolor[rgb]{0.990,0.994,0.992}\phantom{-}0.01 & \cellcolor[rgb]{0.882,0.722,0.739}\textbf{-0.17} & \cellcolor[rgb]{0.881,0.721,0.739}\textbf{-0.014} & \cellcolor[rgb]{0.979,0.950,0.953}-0.002 & \cellcolor[rgb]{0.903,0.771,0.786}\textbf{-0.011} \\
Paraphrasing & \cellcolor[rgb]{0.903,0.772,0.786}\textbf{-0.015} & \cellcolor[rgb]{0.915,0.800,0.813}\textbf{-0.013} & \cellcolor[rgb]{0.988,0.971,0.973}-0.002 & \cellcolor[rgb]{0.845,0.635,0.659}\textbf{-0.22} & \cellcolor[rgb]{0.981,0.956,0.959}-0.03 & \cellcolor[rgb]{0.864,0.679,0.700}-0.19 & \cellcolor[rgb]{0.893,0.748,0.764}\textbf{-0.012} & \cellcolor[rgb]{0.870,0.695,0.715}\textbf{-0.015} & \cellcolor[rgb]{0.947,0.965,0.955}\phantom{-}0.003 \\
\bottomrule
\end{tabular}

%% file: tables/tab_metric_corr.tex
\begin{tabular}{lrrr}
\toprule
 & $\boldsymbol{\Delta p}$ & $\boldsymbol{\Delta\log p}$ & $\boldsymbol{\Phi}$ \\
\midrule
\multicolumn{4}{l}{\emph{Early Answering}} \\
\quad $\Delta p$ & \phantom{-}1.00 & \phantom{-}0.67 & \phantom{-}0.91 \\
\quad $\Delta\log p$ & \phantom{-}0.92 & \phantom{-}1.00 & \phantom{-}0.67 \\
\quad $\Phi$ & \phantom{-}0.79 & \phantom{-}0.80 & \phantom{-}1.00 \\
\addlinespace
\multicolumn{4}{l}{\emph{Filler Tokens}} \\
\quad $\Delta p$ & \phantom{-}1.00 & \phantom{-}0.72 & \phantom{-}0.89 \\
\quad $\Delta\log p$ & \phantom{-}0.98 & \phantom{-}1.00 & \phantom{-}0.71 \\
\quad $\Phi$ & \phantom{-}0.85 & \phantom{-}0.85 & \phantom{-}1.00 \\
\addlinespace
\multicolumn{4}{l}{\emph{Adding Mistakes}} \\
\quad $\Delta p$ & \phantom{-}1.00 & \phantom{-}0.81 & \phantom{-}0.93 \\
\quad $\Delta\log p$ & \phantom{-}1.00 & \phantom{-}1.00 & \phantom{-}0.79 \\
\quad $\Phi$ & \phantom{-}0.48 & \phantom{-}0.48 & \phantom{-}1.00 \\
\addlinespace
\multicolumn{4}{l}{\emph{Paraphrasing}} \\
\quad $\Delta p$ & \phantom{-}1.00 & \phantom{-}0.85 & \phantom{-}0.96 \\
\quad $\Delta\log p$ & \phantom{-}1.00 & \phantom{-}1.00 & \phantom{-}0.84 \\
\quad $\Phi$ & \phantom{-}0.64 & \phantom{-}0.64 & \phantom{-}1.00 \\
\bottomrule
\end{tabular}

%% file: tables/tab_model_agreement_all.tex
\begin{tabular}{lrrrrrr}
\toprule
& \multicolumn{3}{c}{\textbf{Agreement} ($r$)} & \multicolumn{3}{c}{\textbf{Gap} ($\bar\rho$)} \\ \cmidrule(lr){2-4}\cmidrule(l){5-7}
\textbf{Model} & \textbf{Aya} & \textbf{Gemma} & \textbf{Llama} & \textbf{K} & \textbf{R} & \textbf{K}$\,-\,$\textbf{R} \\
\midrule
\multicolumn{7}{l}{\emph{$\Delta p$}} \\
\quad Aya & -- & -- & -- & \cellcolor[rgb]{0.917,0.805,0.818}-0.38 & \cellcolor[rgb]{0.893,0.930,0.910}\phantom{-}0.20 & \cellcolor[rgb]{0.872,0.700,0.719}-0.58 \\
\quad Gemma & \cellcolor[rgb]{0.846,0.899,0.870}0.30 & -- & -- & \cellcolor[rgb]{0.849,0.645,0.667}-0.69 & \cellcolor[rgb]{0.927,0.828,0.839}-0.33 & \cellcolor[rgb]{0.922,0.816,0.828}-0.36 \\
\quad Llama & \cellcolor[rgb]{0.802,0.871,0.834}0.38 & \cellcolor[rgb]{0.669,0.784,0.721}0.63 & -- & \cellcolor[rgb]{0.886,0.731,0.749}-0.52 & \cellcolor[rgb]{0.980,0.954,0.957}-0.09 & \cellcolor[rgb]{0.905,0.778,0.792}-0.43 \\
\quad Qwen & \cellcolor[rgb]{0.817,0.881,0.846}0.35 & \cellcolor[rgb]{0.738,0.829,0.779}0.50 & \cellcolor[rgb]{0.676,0.789,0.727}0.62 & \cellcolor[rgb]{0.960,0.907,0.913}-0.18 & \cellcolor[rgb]{0.979,0.950,0.953}-0.10 & \cellcolor[rgb]{0.982,0.957,0.960}-0.08 \\
\addlinespace
\multicolumn{7}{l}{\emph{$\Delta \log p$}} \\
\quad Aya & -- & -- & -- & \cellcolor[rgb]{0.893,0.748,0.764}-0.49 & \cellcolor[rgb]{0.961,0.975,0.967}\phantom{-}0.07 & \cellcolor[rgb]{0.876,0.709,0.728}-0.56 \\
\quad Gemma & \cellcolor[rgb]{0.944,0.963,0.953}0.11 & -- & -- & \cellcolor[rgb]{0.887,0.735,0.752}-0.51 & \cellcolor[rgb]{0.933,0.842,0.852}-0.31 & \cellcolor[rgb]{0.954,0.892,0.899}-0.21 \\
\quad Llama & \cellcolor[rgb]{0.884,0.924,0.902}0.22 & \cellcolor[rgb]{0.745,0.834,0.786}0.49 & -- & \cellcolor[rgb]{0.921,0.814,0.826}-0.36 & \cellcolor[rgb]{0.994,0.996,0.995}\phantom{-}0.01 & \cellcolor[rgb]{0.918,0.808,0.820}-0.37 \\
\quad Qwen & \cellcolor[rgb]{0.948,0.966,0.956}0.10 & \cellcolor[rgb]{0.809,0.875,0.839}0.37 & \cellcolor[rgb]{0.729,0.823,0.772}0.52 & \cellcolor[rgb]{0.980,0.987,0.983}\phantom{-}0.04 & \cellcolor[rgb]{0.932,0.956,0.943}\phantom{-}0.13 & \cellcolor[rgb]{0.980,0.953,0.956}-0.09 \\
\addlinespace
\multicolumn{7}{l}{\emph{$\Phi$}} \\
\quad Aya & -- & -- & -- & \cellcolor[rgb]{0.932,0.840,0.850}-0.31 & \cellcolor[rgb]{0.892,0.929,0.909}\phantom{-}0.21 & \cellcolor[rgb]{0.887,0.733,0.750}-0.52 \\
\quad Gemma & \cellcolor[rgb]{0.855,0.905,0.878}0.28 & -- & -- & \cellcolor[rgb]{0.844,0.634,0.657}-0.71 & \cellcolor[rgb]{0.924,0.822,0.833}-0.35 & \cellcolor[rgb]{0.920,0.813,0.825}-0.36 \\
\quad Llama & \cellcolor[rgb]{0.853,0.904,0.876}0.28 & \cellcolor[rgb]{0.700,0.804,0.747}0.57 & -- & \cellcolor[rgb]{0.909,0.786,0.800}-0.41 & \cellcolor[rgb]{0.962,0.910,0.915}-0.17 & \cellcolor[rgb]{0.947,0.877,0.884}-0.24 \\
\quad Qwen & \cellcolor[rgb]{0.855,0.905,0.878}0.28 & \cellcolor[rgb]{0.748,0.835,0.787}0.48 & \cellcolor[rgb]{0.690,0.798,0.739}0.59 & \cellcolor[rgb]{0.959,0.904,0.910}-0.19 & \cellcolor[rgb]{0.957,0.898,0.905}-0.20 & \cellcolor[rgb]{0.994,0.996,0.995}\phantom{-}0.01 \\
\bottomrule
\end{tabular}

%% file: tables/tab_mediation_all.tex
\begin{tabular}{lrrrrr}
\toprule
\multicolumn{6}{l}{$\boldsymbol{\Delta p}$} \\
& \textbf{K}\,$\cdot$\,\textbf{Mistakes} & \textbf{K}\,$\cdot$\,\textbf{Paraphrasing} & \textbf{R}\,$\cdot$\,\textbf{Early} & \textbf{R}\,$\cdot$\,\textbf{Paraphrasing} \\
& $\beta_1 = -0.51$ & $\beta_1 = -0.34$ & $\beta_1 = 0.34$ & $\beta_1 = -0.30$ \\
\midrule
\multicolumn{5}{l}{\emph{Final answer}} \\
\quad Probability & \cellcolor[rgb]{0.910,0.941,0.924}$17$ & \cellcolor[rgb]{0.745,0.833,0.785}$49$ & \cellcolor[rgb]{0.899,0.934,0.915}$19$ & \cellcolor[rgb]{0.979,0.952,0.955}$-9$ \\
\quad Log-probability & \cellcolor[rgb]{0.943,0.963,0.952}$11$ & \cellcolor[rgb]{0.803,0.872,0.834}$38$ & \cellcolor[rgb]{0.928,0.953,0.939}$14$ & \cellcolor[rgb]{0.999,0.999,0.999}$0$ \\
\quad Entropy & \cellcolor[rgb]{0.935,0.847,0.856}$-30$ & \cellcolor[rgb]{0.973,0.937,0.941}$-12$ & \cellcolor[rgb]{0.890,0.928,0.908}$21$ & \cellcolor[rgb]{0.945,0.871,0.879}$-25$ \\
\quad Top-1 probability & \cellcolor[rgb]{0.939,0.858,0.867}$-28$ & \cellcolor[rgb]{0.986,0.967,0.969}$-6$ & \cellcolor[rgb]{0.880,0.922,0.899}$23$ & \cellcolor[rgb]{0.942,0.864,0.873}$-26$ \\
\quad Margin & \cellcolor[rgb]{0.942,0.863,0.872}$-27$ & \cellcolor[rgb]{0.998,0.996,0.997}$-1$ & \cellcolor[rgb]{0.878,0.920,0.897}$23$ & \cellcolor[rgb]{0.944,0.867,0.876}$-26$ \\
\addlinespace
\multicolumn{5}{l}{\emph{Reasoning chain}} \\
\quad Log-probability & \cellcolor[rgb]{0.991,0.994,0.993}$2$ & \cellcolor[rgb]{0.990,0.994,0.992}$2$ & \cellcolor[rgb]{0.984,0.990,0.987}$3$ & \cellcolor[rgb]{0.907,0.940,0.922}$18$ \\
\quad Entropy & \cellcolor[rgb]{0.981,0.988,0.984}$4$ & \cellcolor[rgb]{0.997,0.992,0.993}$-1$ & \cellcolor[rgb]{0.870,0.915,0.891}$25$ & \cellcolor[rgb]{0.960,0.974,0.966}$8$ \\
\quad Top-1 probability & \cellcolor[rgb]{0.994,0.987,0.988}$-3$ & \cellcolor[rgb]{0.993,0.996,0.994}$1$ & \cellcolor[rgb]{0.911,0.942,0.925}$17$ & \cellcolor[rgb]{0.989,0.993,0.991}$2$ \\
\quad Margin & \cellcolor[rgb]{0.982,0.958,0.961}$-8$ & \cellcolor[rgb]{0.979,0.986,0.982}$4$ & \cellcolor[rgb]{0.952,0.968,0.959}$9$ & \cellcolor[rgb]{0.996,0.998,0.997}$1$ \\
\midrule \multicolumn{6}{l}{$\boldsymbol{\Delta \log p}$} \\
& \textbf{K}\,$\cdot$\,\textbf{Filler} & \textbf{K}\,$\cdot$\,\textbf{Mistakes} & \textbf{K}\,$\cdot$\,\textbf{Paraphrasing} & \textbf{R}\,$\cdot$\,\textbf{Early} \\
& $\beta_1 = -0.21$ & $\beta_1 = -0.42$ & $\beta_1 = -0.26$ & $\beta_1 = 0.21$ \\
\midrule \multicolumn{5}{l}{\emph{Final answer}} \\
\quad Probability & \cellcolor[rgb]{0.478,0.659,0.560}$125$ & \cellcolor[rgb]{0.543,0.702,0.615}$87$ & \cellcolor[rgb]{0.478,0.659,0.560}$141$ & \cellcolor[rgb]{0.972,0.933,0.938}$-13$ \\
\quad Log-probability & \cellcolor[rgb]{0.488,0.666,0.569}$98$ & \cellcolor[rgb]{0.636,0.762,0.693}$70$ & \cellcolor[rgb]{0.478,0.659,0.560}$109$ & \cellcolor[rgb]{0.968,0.924,0.929}$-15$ \\
\quad Entropy & \cellcolor[rgb]{0.978,0.949,0.952}$-10$ & \cellcolor[rgb]{0.967,0.923,0.928}$-15$ & \cellcolor[rgb]{0.952,0.888,0.895}$-22$ & \cellcolor[rgb]{0.929,0.954,0.941}$14$ \\
\quad Top-1 probability & \cellcolor[rgb]{0.980,0.952,0.955}$-9$ & \cellcolor[rgb]{0.974,0.939,0.942}$-12$ & \cellcolor[rgb]{0.956,0.897,0.904}$-20$ & \cellcolor[rgb]{0.930,0.954,0.941}$13$ \\
\quad Margin & \cellcolor[rgb]{0.998,0.994,0.995}$-1$ & \cellcolor[rgb]{0.985,0.965,0.967}$-7$ & \cellcolor[rgb]{0.978,0.949,0.952}$-10$ & \cellcolor[rgb]{0.935,0.957,0.945}$13$ \\
\addlinespace
\multicolumn{5}{l}{\emph{Reasoning chain}} \\
\quad Log-probability & \cellcolor[rgb]{0.982,0.988,0.985}$4$ & \cellcolor[rgb]{0.991,0.994,0.993}$2$ & \cellcolor[rgb]{0.985,0.990,0.987}$3$ & \cellcolor[rgb]{0.952,0.969,0.959}$9$ \\
\quad Entropy & \cellcolor[rgb]{0.956,0.971,0.963}$8$ & \cellcolor[rgb]{0.975,0.984,0.979}$5$ & \cellcolor[rgb]{0.992,0.995,0.994}$1$ & \cellcolor[rgb]{0.899,0.934,0.915}$19$ \\
\quad Top-1 probability & \cellcolor[rgb]{0.989,0.975,0.977}$-5$ & \cellcolor[rgb]{0.992,0.982,0.983}$-3$ & \cellcolor[rgb]{0.997,0.992,0.993}$-2$ & \cellcolor[rgb]{0.933,0.956,0.943}$13$ \\
\quad Margin & \cellcolor[rgb]{0.969,0.926,0.931}$-14$ & \cellcolor[rgb]{0.976,0.944,0.948}$-11$ & \cellcolor[rgb]{0.988,0.971,0.973}$-6$ & \cellcolor[rgb]{0.965,0.977,0.971}$7$ \\
\midrule \multicolumn{6}{l}{$\boldsymbol{\Phi}$} \\
& \textbf{K}\,$\cdot$\,\textbf{Mistakes} & \textbf{K}\,$\cdot$\,\textbf{Paraphrasing} & \textbf{R}\,$\cdot$\,\textbf{Early} & \textbf{R}\,$\cdot$\,\textbf{Filler} & \textbf{R}\,$\cdot$\,\textbf{Paraphrasing} \\
& $\beta_1 = -0.39$ & $\beta_1 = -0.27$ & $\beta_1 = 0.19$ & $\beta_1 = 0.31$ & $\beta_1 = -0.35$ \\
\midrule \multicolumn{6}{l}{\emph{Final answer}} \\
\quad Probability & \cellcolor[rgb]{0.931,0.838,0.848}$-31$ & \cellcolor[rgb]{0.812,0.877,0.842}$36$ & \cellcolor[rgb]{0.937,0.851,0.860}$-29$ & \cellcolor[rgb]{0.896,0.756,0.772}$-47$ & \cellcolor[rgb]{0.988,0.992,0.990}$2$ \\
\quad Log-probability & \cellcolor[rgb]{0.944,0.868,0.876}$-26$ & \cellcolor[rgb]{0.823,0.884,0.851}$34$ & \cellcolor[rgb]{0.939,0.856,0.866}$-28$ & \cellcolor[rgb]{0.890,0.741,0.757}$-50$ & \cellcolor[rgb]{0.966,0.978,0.971}$6$ \\
\quad Entropy & \cellcolor[rgb]{0.917,0.805,0.818}$-38$ & \cellcolor[rgb]{0.875,0.918,0.895}$24$ & \cellcolor[rgb]{0.972,0.934,0.939}$-13$ & \cellcolor[rgb]{0.964,0.914,0.920}$-17$ & \cellcolor[rgb]{0.988,0.973,0.974}$-5$ \\
\quad Top-1 probability & \cellcolor[rgb]{0.925,0.824,0.835}$-34$ & \cellcolor[rgb]{0.809,0.875,0.839}$37$ & \cellcolor[rgb]{0.960,0.907,0.913}$-18$ & \cellcolor[rgb]{0.923,0.819,0.830}$-35$ & \cellcolor[rgb]{0.999,0.997,0.997}$-1$ \\
\quad Margin & \cellcolor[rgb]{0.922,0.817,0.829}$-35$ & \cellcolor[rgb]{0.786,0.860,0.819}$41$ & \cellcolor[rgb]{0.954,0.892,0.899}$-21$ & \cellcolor[rgb]{0.907,0.781,0.795}$-42$ & \cellcolor[rgb]{0.987,0.992,0.989}$2$ \\
\addlinespace
\multicolumn{6}{l}{\emph{Reasoning chain}} \\
\quad Log-probability & \cellcolor[rgb]{0.995,0.997,0.996}$1$ & \cellcolor[rgb]{0.999,0.999,0.999}$0$ & \cellcolor[rgb]{0.936,0.958,0.946}$12$ & \cellcolor[rgb]{0.941,0.962,0.951}$11$ & \cellcolor[rgb]{0.959,0.973,0.965}$8$ \\
\quad Entropy & \cellcolor[rgb]{0.981,0.987,0.984}$4$ & \cellcolor[rgb]{0.990,0.975,0.977}$-5$ & \cellcolor[rgb]{0.771,0.851,0.807}$44$ & \cellcolor[rgb]{0.837,0.894,0.863}$31$ & \cellcolor[rgb]{0.999,0.999,0.999}$0$ \\
\quad Top-1 probability & \cellcolor[rgb]{0.993,0.984,0.985}$-3$ & \cellcolor[rgb]{0.974,0.983,0.978}$5$ & \cellcolor[rgb]{0.846,0.900,0.871}$29$ & \cellcolor[rgb]{0.878,0.920,0.897}$23$ & \cellcolor[rgb]{0.994,0.986,0.987}$-3$ \\
\quad Margin & \cellcolor[rgb]{0.981,0.955,0.958}$-9$ & \cellcolor[rgb]{0.923,0.950,0.935}$15$ & \cellcolor[rgb]{0.917,0.946,0.931}$16$ & \cellcolor[rgb]{0.926,0.952,0.938}$14$ & \cellcolor[rgb]{0.996,0.990,0.990}$-2$ \\
\bottomrule
\end{tabular}

%% file: tables/tab_confidence_gradient.tex
\begin{tabular}{lrrr}
\toprule
\textbf{Confidence} & $\boldsymbol{\hat\beta^{\mathrm{K}}}$ & $\boldsymbol{\hat\beta^{\mathrm{R}}}$ & $\boldsymbol{\hat\beta^{\mathrm{K}} - \hat\beta^{\mathrm{R}}}$ \\
\midrule
\multicolumn{4}{l}{\emph{Final answer}} \\
\quad Probability & \cellcolor[rgb]{0.478,0.659,0.560}\textbf{\phantom{-}0.24} & \cellcolor[rgb]{0.511,0.681,0.588}\textbf{\phantom{-}0.23} & \cellcolor[rgb]{0.967,0.978,0.972}\phantom{-}0.02 \\
\quad Log-probability & \cellcolor[rgb]{0.506,0.677,0.584}\textbf{\phantom{-}0.23} & \cellcolor[rgb]{0.511,0.681,0.588}\textbf{\phantom{-}0.23} & \cellcolor[rgb]{0.995,0.997,0.996}\phantom{-}0.00 \\
\quad Entropy & \cellcolor[rgb]{0.797,0.522,0.552}\textbf{-0.22} & \cellcolor[rgb]{0.815,0.565,0.593}\textbf{-0.20} & \cellcolor[rgb]{0.982,0.957,0.960}\textbf{-0.02} \\
\quad Top-1 probability & \cellcolor[rgb]{0.542,0.701,0.614}\textbf{\phantom{-}0.21} & \cellcolor[rgb]{0.505,0.677,0.583}\textbf{\phantom{-}0.23} & \cellcolor[rgb]{0.985,0.964,0.966}-0.02 \\
\quad Margin & \cellcolor[rgb]{0.546,0.704,0.618}\textbf{\phantom{-}0.21} & \cellcolor[rgb]{0.489,0.667,0.570}\textbf{\phantom{-}0.24} & \cellcolor[rgb]{0.976,0.943,0.947}-0.03 \\
\addlinespace
\multicolumn{4}{l}{\emph{Reasoning chain}} \\
\quad Log-probability & \cellcolor[rgb]{0.970,0.980,0.974}\phantom{-}0.01 & \cellcolor[rgb]{0.789,0.862,0.822}\phantom{-}0.10 & \cellcolor[rgb]{0.924,0.821,0.833}-0.08 \\
\quad Entropy & \cellcolor[rgb]{0.941,0.861,0.870}-0.07 & \cellcolor[rgb]{0.857,0.665,0.686}-0.16 & \cellcolor[rgb]{0.802,0.871,0.833}\phantom{-}0.09 \\
\quad Top-1 probability & \cellcolor[rgb]{0.970,0.930,0.935}-0.03 & \cellcolor[rgb]{0.757,0.841,0.795}\phantom{-}0.11 & \cellcolor[rgb]{0.868,0.690,0.709}\textbf{-0.15} \\
\quad Margin & \cellcolor[rgb]{0.910,0.788,0.802}-0.10 & \cellcolor[rgb]{0.848,0.901,0.872}\phantom{-}0.07 & \cellcolor[rgb]{0.846,0.637,0.660}\textbf{-0.17} \\
\bottomrule
\end{tabular}

%% file: tables/tab_simpson_summary.tex
\begin{tabular}{lrrrr}
\toprule
& $\boldsymbol{\overline{|r_{\mathrm{b}}|}}$ & $\boldsymbol{\overline{|r_{\mathrm{w}}|}}$ & $\boldsymbol{|r_{\mathrm{w}}| < .1}$ & \textbf{Rev.} \\
\midrule
\multicolumn{5}{l}{\emph{$\Delta p$}} \\
\quad Answer\,$\cdot$\,K & $0.60$ & $0.05$ & $17/20$ & $10/20$ \\
\quad Answer\,$\cdot$\,R & $0.59$ & $0.11$ & $10/20$ & $5/20$ \\
\quad Chain\,$\cdot$\,K & $0.27$ & $0.02$ & $16/16$ & $6/16$ \\
\quad Chain\,$\cdot$\,R & $0.41$ & $0.06$ & $12/16$ & $5/16$ \\
\addlinespace
\multicolumn{5}{l}{\emph{$\Delta \log p$}} \\
\quad Answer\,$\cdot$\,K & $0.64$ & $0.04$ & $19/20$ & $1/20$ \\
\quad Answer\,$\cdot$\,R & $0.40$ & $0.05$ & $17/20$ & $12/20$ \\
\quad Chain\,$\cdot$\,K & $0.24$ & $0.08$ & $9/16$ & $8/16$ \\
\quad Chain\,$\cdot$\,R & $0.40$ & $0.09$ & $10/16$ & $8/16$ \\
\addlinespace
\multicolumn{5}{l}{\emph{$\Phi$}} \\
\quad Answer\,$\cdot$\,K & $0.55$ & $0.10$ & $11/20$ & $10/20$ \\
\quad Answer\,$\cdot$\,R & $0.64$ & $0.09$ & $15/20$ & $15/20$ \\
\quad Chain\,$\cdot$\,K & $0.28$ & $0.02$ & $16/16$ & $10/16$ \\
\quad Chain\,$\cdot$\,R & $0.41$ & $0.06$ & $12/16$ & $3/16$ \\
\addlinespace
\textbf{All} & $0.47$ & $0.06$ & $164/216$ & $93/216$ \\
\bottomrule
\end{tabular}

%% file: tables/tab_rate_magnitude_all.tex
\begin{tabular}{lrrr}
\toprule
\textbf{Intervention} & $\boldsymbol{\hat\beta^{\mathrm{K}}}$ & $\boldsymbol{\hat\beta^{\mathrm{R}}}$ & $\boldsymbol{\hat\beta^{\mathrm{K}} - \hat\beta^{\mathrm{R}}}$ \\
\midrule
\multicolumn{4}{l}{\emph{Early Answering}} \\
\quad Reliance & \cellcolor[rgb]{0.946,0.874,0.882}-0.08 & \cellcolor[rgb]{0.605,0.742,0.667}\textbf{\phantom{-}0.24} & \cellcolor[rgb]{0.780,0.483,0.516}\textbf{-0.31} \\
\quad Rate & \cellcolor[rgb]{0.655,0.775,0.710}\phantom{-}0.003 & \cellcolor[rgb]{0.478,0.659,0.560}\phantom{-}0.004 & \cellcolor[rgb]{0.925,0.824,0.835}-0.001 \\
\quad Magnitude & \cellcolor[rgb]{0.989,0.975,0.977}-0.01 & \cellcolor[rgb]{0.503,0.676,0.581}\textbf{\phantom{-}0.24} & \cellcolor[rgb]{0.780,0.483,0.516}\textbf{-0.25} \\
\addlinespace
\multicolumn{4}{l}{\emph{Filler Tokens}} \\
\quad Reliance & \cellcolor[rgb]{0.861,0.673,0.693}\textbf{-0.15} & \cellcolor[rgb]{0.808,0.875,0.839}\phantom{-}0.08 & \cellcolor[rgb]{0.780,0.483,0.516}\textbf{-0.23} \\
\quad Rate & \cellcolor[rgb]{0.780,0.483,0.516}-0.030 & \cellcolor[rgb]{0.994,0.986,0.987}-0.001 & \cellcolor[rgb]{0.786,0.497,0.529}\textbf{-0.029} \\
\quad Magnitude & \cellcolor[rgb]{0.975,0.984,0.979}\phantom{-}0.00 & \cellcolor[rgb]{0.478,0.659,0.560}\phantom{-}0.10 & \cellcolor[rgb]{0.791,0.508,0.539}-0.10 \\
\addlinespace
\multicolumn{4}{l}{\emph{Adding Mistakes}} \\
\quad Reliance & \cellcolor[rgb]{0.788,0.501,0.533}\textbf{-0.16} & \cellcolor[rgb]{0.982,0.988,0.985}\phantom{-}0.01 & \cellcolor[rgb]{0.780,0.483,0.516}\textbf{-0.17} \\
\quad Rate & \cellcolor[rgb]{0.780,0.483,0.516}\textbf{-0.044} & \cellcolor[rgb]{0.894,0.750,0.766}\textbf{-0.021} & \cellcolor[rgb]{0.886,0.733,0.750}\textbf{-0.023} \\
\quad Magnitude & \cellcolor[rgb]{0.823,0.884,0.851}\phantom{-}0.10 & \cellcolor[rgb]{0.478,0.659,0.560}\textbf{\phantom{-}0.29} & \cellcolor[rgb]{0.855,0.658,0.680}-0.19 \\
\addlinespace
\multicolumn{4}{l}{\emph{Paraphrasing}} \\
\quad Reliance & \cellcolor[rgb]{0.780,0.483,0.516}\textbf{-0.22} & \cellcolor[rgb]{0.973,0.938,0.942}-0.03 & \cellcolor[rgb]{0.807,0.545,0.574}-0.19 \\
\quad Rate & \cellcolor[rgb]{0.780,0.483,0.516}\textbf{-0.047} & \cellcolor[rgb]{0.819,0.575,0.603}\textbf{-0.039} & \cellcolor[rgb]{0.961,0.908,0.913}\textbf{-0.008} \\
\quad Magnitude & \cellcolor[rgb]{0.478,0.659,0.560}\textbf{\phantom{-}0.64} & \cellcolor[rgb]{0.554,0.709,0.624}\textbf{\phantom{-}0.55} & \cellcolor[rgb]{0.924,0.950,0.936}\phantom{-}0.09 \\
\bottomrule
\end{tabular}

%% file: tables/tab_robustness_all.tex
\begin{tabular}{lrrrrrrrrr}
\toprule
 & \multicolumn{3}{c}{$\boldsymbol{\Delta p}$} & \multicolumn{3}{c}{$\boldsymbol{\Delta\log p}$} & \multicolumn{3}{c}{$\boldsymbol{\Phi}$} \\ \cmidrule(lr){2-4} \cmidrule(lr){5-7} \cmidrule(lr){8-10}
\textbf{Specification} & $\boldsymbol{\mathrm{K}}$ & $\boldsymbol{\mathrm{R}}$ & $\boldsymbol{\mathrm{K}-\mathrm{R}}$ & $\boldsymbol{\mathrm{K}}$ & $\boldsymbol{\mathrm{R}}$ & $\boldsymbol{\mathrm{K}-\mathrm{R}}$ & $\boldsymbol{\mathrm{K}}$ & $\boldsymbol{\mathrm{R}}$ & $\boldsymbol{\mathrm{K}-\mathrm{R}}$ \\
\midrule
\multicolumn{10}{l}{\emph{Early Answering}} \\
\quad Primary specification & \cellcolor[rgb]{0.784,0.859,0.818}\phantom{-}0.015 & \cellcolor[rgb]{0.656,0.775,0.710}\textbf{\phantom{-}0.024} & \cellcolor[rgb]{0.946,0.873,0.881}-0.009 & \cellcolor[rgb]{0.949,0.881,0.889}-0.08 & \cellcolor[rgb]{0.628,0.757,0.686}\textbf{\phantom{-}0.24} & \cellcolor[rgb]{0.793,0.513,0.544}\textbf{-0.31} & \cellcolor[rgb]{0.794,0.865,0.826}\phantom{-}0.013 & \cellcolor[rgb]{0.765,0.847,0.802}\textbf{\phantom{-}0.014} & \cellcolor[rgb]{0.988,0.972,0.974}-0.002 \\
\quad \quad + Item fixed effects & \cellcolor[rgb]{0.865,0.912,0.886}\phantom{-}0.009 & \cellcolor[rgb]{0.714,0.814,0.759}\textbf{\phantom{-}0.020} & \cellcolor[rgb]{0.937,0.851,0.861}-0.011 & \cellcolor[rgb]{0.934,0.845,0.855}-0.10 & \cellcolor[rgb]{0.678,0.790,0.729}\textbf{\phantom{-}0.20} & \cellcolor[rgb]{0.798,0.526,0.557}\textbf{-0.30} & \cellcolor[rgb]{0.852,0.903,0.875}\phantom{-}0.009 & \cellcolor[rgb]{0.821,0.883,0.849}\textbf{\phantom{-}0.011} & \cellcolor[rgb]{0.987,0.970,0.972}-0.002 \\
\quad \quad + CoT length & \cellcolor[rgb]{0.811,0.877,0.841}\phantom{-}0.013 & \cellcolor[rgb]{0.685,0.794,0.735}\textbf{\phantom{-}0.022} & \cellcolor[rgb]{0.947,0.875,0.883}-0.009 & \cellcolor[rgb]{0.931,0.839,0.849}-0.10 & \cellcolor[rgb]{0.640,0.765,0.697}\textbf{\phantom{-}0.23} & \cellcolor[rgb]{0.780,0.483,0.516}\textbf{-0.33} & \cellcolor[rgb]{0.823,0.884,0.851}\phantom{-}0.011 & \cellcolor[rgb]{0.793,0.865,0.826}\textbf{\phantom{-}0.013} & \cellcolor[rgb]{0.987,0.971,0.972}-0.002 \\
\quad \quad + Answer confidence & \cellcolor[rgb]{0.685,0.794,0.735}\textbf{\phantom{-}0.022} & \cellcolor[rgb]{0.620,0.752,0.680}\textbf{\phantom{-}0.027} & \cellcolor[rgb]{0.973,0.936,0.940}-0.005 & \cellcolor[rgb]{0.952,0.969,0.960}\phantom{-}0.03 & \cellcolor[rgb]{0.505,0.677,0.583}\textbf{\phantom{-}0.31} & \cellcolor[rgb]{0.812,0.557,0.586}\textbf{-0.28} & \cellcolor[rgb]{0.576,0.723,0.643}\textbf{\phantom{-}0.026} & \cellcolor[rgb]{0.620,0.752,0.680}\textbf{\phantom{-}0.023} & \cellcolor[rgb]{0.956,0.971,0.963}\phantom{-}0.003 \\
\addlinespace[2pt]
\quad Correct answers only & \cellcolor[rgb]{0.803,0.871,0.834}\phantom{-}0.014 & \cellcolor[rgb]{0.599,0.738,0.662}\phantom{-}0.028 & \cellcolor[rgb]{0.914,0.798,0.811}-0.014 & \cellcolor[rgb]{0.917,0.946,0.930}\phantom{-}0.05 & \cellcolor[rgb]{0.505,0.677,0.584}\textbf{\phantom{-}0.31} & \cellcolor[rgb]{0.827,0.592,0.618}\textbf{-0.26} & \cellcolor[rgb]{0.852,0.903,0.875}\phantom{-}0.009 & \cellcolor[rgb]{0.625,0.756,0.685}\phantom{-}0.023 & \cellcolor[rgb]{0.905,0.776,0.790}-0.014 \\
\quad Cells with ${\geq}\,30$ generations & \cellcolor[rgb]{0.779,0.856,0.814}\phantom{-}0.016 & \cellcolor[rgb]{0.785,0.859,0.819}\textbf{\phantom{-}0.015} & \cellcolor[rgb]{0.994,0.996,0.995}\phantom{-}0.000 & \cellcolor[rgb]{0.874,0.918,0.894}\phantom{-}0.08 & \cellcolor[rgb]{0.839,0.895,0.865}\phantom{-}0.10 & \cellcolor[rgb]{0.985,0.965,0.967}-0.02 & \cellcolor[rgb]{0.798,0.868,0.830}\phantom{-}0.012 & \cellcolor[rgb]{0.866,0.912,0.887}\phantom{-}0.008 & \cellcolor[rgb]{0.932,0.956,0.943}\phantom{-}0.004 \\
\quad Without Aya and Llama & \cellcolor[rgb]{0.647,0.770,0.703}\phantom{-}0.025 & \cellcolor[rgb]{0.745,0.834,0.786}\textbf{\phantom{-}0.018} & \cellcolor[rgb]{0.902,0.936,0.917}\phantom{-}0.007 & \cellcolor[rgb]{0.721,0.818,0.765}\textbf{\phantom{-}0.18} & \cellcolor[rgb]{0.830,0.889,0.857}\phantom{-}0.11 & \cellcolor[rgb]{0.891,0.929,0.908}\phantom{-}0.07 & \cellcolor[rgb]{0.640,0.765,0.697}\phantom{-}0.022 & \cellcolor[rgb]{0.823,0.885,0.851}\phantom{-}0.011 & \cellcolor[rgb]{0.816,0.880,0.845}\phantom{-}0.011 \\
\addlinespace
\multicolumn{10}{l}{\emph{Filler Tokens}} \\
\quad Primary specification & \cellcolor[rgb]{0.917,0.806,0.818}-0.014 & \cellcolor[rgb]{0.707,0.809,0.753}\phantom{-}0.021 & \cellcolor[rgb]{0.794,0.516,0.547}\textbf{-0.034} & \cellcolor[rgb]{0.903,0.771,0.786}\textbf{-0.15} & \cellcolor[rgb]{0.866,0.913,0.887}\phantom{-}0.08 & \cellcolor[rgb]{0.846,0.639,0.662}\textbf{-0.23} & \cellcolor[rgb]{0.983,0.961,0.963}-0.002 & \cellcolor[rgb]{0.625,0.755,0.684}\textbf{\phantom{-}0.023} & \cellcolor[rgb]{0.825,0.590,0.616}\textbf{-0.026} \\
\quad \quad + Item fixed effects & \cellcolor[rgb]{0.915,0.801,0.814}-0.014 & \cellcolor[rgb]{0.788,0.862,0.821}\phantom{-}0.015 & \cellcolor[rgb]{0.826,0.591,0.617}\textbf{-0.029} & \cellcolor[rgb]{0.907,0.781,0.795}\textbf{-0.14} & \cellcolor[rgb]{0.913,0.943,0.926}\phantom{-}0.06 & \cellcolor[rgb]{0.870,0.695,0.714}\textbf{-0.20} & \cellcolor[rgb]{0.989,0.973,0.975}-0.002 & \cellcolor[rgb]{0.733,0.826,0.775}\phantom{-}0.016 & \cellcolor[rgb]{0.876,0.709,0.728}\textbf{-0.018} \\
\quad \quad + CoT length & \cellcolor[rgb]{0.934,0.844,0.854}-0.011 & \cellcolor[rgb]{0.726,0.821,0.769}\textbf{\phantom{-}0.019} & \cellcolor[rgb]{0.818,0.572,0.600}\textbf{-0.030} & \cellcolor[rgb]{0.899,0.762,0.777}\textbf{-0.15} & \cellcolor[rgb]{0.881,0.922,0.900}\phantom{-}0.08 & \cellcolor[rgb]{0.848,0.644,0.666}\textbf{-0.23} & \cellcolor[rgb]{0.995,0.989,0.990}-0.001 & \cellcolor[rgb]{0.640,0.765,0.697}\textbf{\phantom{-}0.022} & \cellcolor[rgb]{0.844,0.633,0.656}\textbf{-0.023} \\
\quad \quad + Answer confidence & \cellcolor[rgb]{0.988,0.971,0.973}-0.002 & \cellcolor[rgb]{0.628,0.757,0.687}\textbf{\phantom{-}0.026} & \cellcolor[rgb]{0.831,0.603,0.628}\textbf{-0.028} & \cellcolor[rgb]{0.925,0.825,0.836}\textbf{-0.11} & \cellcolor[rgb]{0.826,0.887,0.854}\phantom{-}0.11 & \cellcolor[rgb]{0.852,0.653,0.675}\textbf{-0.22} & \cellcolor[rgb]{0.841,0.896,0.866}\phantom{-}0.010 & \cellcolor[rgb]{0.502,0.675,0.580}\textbf{\phantom{-}0.031} & \cellcolor[rgb]{0.857,0.665,0.686}\textbf{-0.021} \\
\addlinespace[2pt]
\quad Correct answers only & \cellcolor[rgb]{0.897,0.758,0.773}-0.017 & \cellcolor[rgb]{0.723,0.819,0.766}\phantom{-}0.019 & \cellcolor[rgb]{0.780,0.483,0.516}\textbf{-0.037} & \cellcolor[rgb]{0.959,0.904,0.910}-0.06 & \cellcolor[rgb]{0.771,0.851,0.807}\phantom{-}0.14 & \cellcolor[rgb]{0.863,0.677,0.698}\textbf{-0.21} & \cellcolor[rgb]{0.930,0.834,0.845}-0.010 & \cellcolor[rgb]{0.645,0.768,0.701}\phantom{-}0.022 & \cellcolor[rgb]{0.780,0.483,0.516}\textbf{-0.032} \\
\quad Cells with ${\geq}\,30$ generations & \cellcolor[rgb]{0.905,0.776,0.790}-0.016 & \cellcolor[rgb]{0.922,0.949,0.935}\phantom{-}0.005 & \cellcolor[rgb]{0.872,0.699,0.718}\textbf{-0.021} & \cellcolor[rgb]{0.978,0.949,0.952}-0.03 & \cellcolor[rgb]{0.985,0.964,0.966}-0.02 & \cellcolor[rgb]{0.993,0.985,0.986}-0.01 & \cellcolor[rgb]{0.900,0.764,0.779}-0.015 & \cellcolor[rgb]{0.899,0.934,0.915}\phantom{-}0.006 & \cellcolor[rgb]{0.857,0.665,0.686}\textbf{-0.021} \\
\quad Without Aya and Llama & \cellcolor[rgb]{0.927,0.829,0.840}-0.012 & \cellcolor[rgb]{0.867,0.913,0.888}\phantom{-}0.009 & \cellcolor[rgb]{0.871,0.698,0.717}\textbf{-0.021} & \cellcolor[rgb]{0.995,0.989,0.990}-0.01 & \cellcolor[rgb]{0.916,0.945,0.929}\phantom{-}0.05 & \cellcolor[rgb]{0.960,0.906,0.912}-0.06 & \cellcolor[rgb]{0.916,0.802,0.815}-0.012 & \cellcolor[rgb]{0.927,0.952,0.938}\phantom{-}0.005 & \cellcolor[rgb]{0.885,0.730,0.747}\textbf{-0.017} \\
\addlinespace
\multicolumn{10}{l}{\emph{Adding Mistakes}} \\
\quad Primary specification & \cellcolor[rgb]{0.881,0.720,0.738}\textbf{-0.020} & \cellcolor[rgb]{0.979,0.950,0.953}-0.004 & \cellcolor[rgb]{0.902,0.770,0.785}\textbf{-0.016} & \cellcolor[rgb]{0.892,0.747,0.763}\textbf{-0.16} & \cellcolor[rgb]{0.991,0.994,0.992}\phantom{-}0.01 & \cellcolor[rgb]{0.888,0.737,0.754}\textbf{-0.17} & \cellcolor[rgb]{0.906,0.779,0.793}\textbf{-0.014} & \cellcolor[rgb]{0.983,0.960,0.963}-0.002 & \cellcolor[rgb]{0.923,0.818,0.830}\textbf{-0.011} \\
\quad \quad + Item fixed effects & \cellcolor[rgb]{0.898,0.761,0.776}\textbf{-0.017} & \cellcolor[rgb]{0.964,0.916,0.921}-0.006 & \cellcolor[rgb]{0.934,0.845,0.855}\textbf{-0.011} & \cellcolor[rgb]{0.909,0.787,0.800}\textbf{-0.14} & \cellcolor[rgb]{0.984,0.963,0.965}-0.02 & \cellcolor[rgb]{0.925,0.824,0.835}\textbf{-0.11} & \cellcolor[rgb]{0.914,0.798,0.811}\textbf{-0.013} & \cellcolor[rgb]{0.967,0.921,0.926}-0.005 & \cellcolor[rgb]{0.948,0.877,0.885}\textbf{-0.008} \\
\quad \quad + CoT length & \cellcolor[rgb]{0.889,0.739,0.756}\textbf{-0.018} & \cellcolor[rgb]{0.977,0.945,0.949}-0.004 & \cellcolor[rgb]{0.912,0.794,0.807}\textbf{-0.015} & \cellcolor[rgb]{0.895,0.752,0.768}\textbf{-0.16} & \cellcolor[rgb]{0.995,0.996,0.995}\phantom{-}0.00 & \cellcolor[rgb]{0.892,0.747,0.763}\textbf{-0.16} & \cellcolor[rgb]{0.914,0.798,0.810}\textbf{-0.013} & \cellcolor[rgb]{0.982,0.957,0.960}-0.003 & \cellcolor[rgb]{0.932,0.841,0.851}\textbf{-0.010} \\
\quad \quad + Perturber similarity & \cellcolor[rgb]{0.979,0.950,0.953}-0.004 & \cellcolor[rgb]{0.935,0.958,0.945}\phantom{-}0.005 & \cellcolor[rgb]{0.951,0.886,0.893}\textbf{-0.008} & \cellcolor[rgb]{0.949,0.880,0.887}\textbf{-0.08} & \cellcolor[rgb]{0.923,0.949,0.935}\phantom{-}0.05 & \cellcolor[rgb]{0.916,0.803,0.816}\textbf{-0.13} & \cellcolor[rgb]{0.988,0.973,0.975}-0.002 & \cellcolor[rgb]{0.919,0.947,0.932}\phantom{-}0.005 & \cellcolor[rgb]{0.954,0.893,0.900}-0.007 \\
\quad \quad + Answer confidence & \cellcolor[rgb]{0.916,0.803,0.815}\textbf{-0.014} & \cellcolor[rgb]{0.994,0.996,0.995}\phantom{-}0.000 & \cellcolor[rgb]{0.914,0.797,0.810}\textbf{-0.014} & \cellcolor[rgb]{0.909,0.785,0.799}\textbf{-0.14} & \cellcolor[rgb]{0.964,0.977,0.970}\phantom{-}0.02 & \cellcolor[rgb]{0.893,0.749,0.765}\textbf{-0.16} & \cellcolor[rgb]{0.956,0.898,0.904}-0.006 & \cellcolor[rgb]{0.945,0.964,0.953}\phantom{-}0.003 & \cellcolor[rgb]{0.933,0.843,0.853}\textbf{-0.010} \\
\addlinespace[2pt]
\quad Correct answers only & \cellcolor[rgb]{0.936,0.850,0.859}-0.011 & \cellcolor[rgb]{0.990,0.977,0.978}-0.002 & \cellcolor[rgb]{0.946,0.873,0.881}-0.009 & \cellcolor[rgb]{0.950,0.882,0.890}-0.08 & \cellcolor[rgb]{0.963,0.976,0.969}\phantom{-}0.02 & \cellcolor[rgb]{0.934,0.846,0.856}\textbf{-0.10} & \cellcolor[rgb]{0.933,0.842,0.852}-0.010 & \cellcolor[rgb]{0.997,0.993,0.994}-0.000 & \cellcolor[rgb]{0.936,0.849,0.859}\textbf{-0.009} \\
\quad Single-word edits only & \cellcolor[rgb]{0.951,0.885,0.892}\textbf{-0.008} & \cellcolor[rgb]{0.974,0.938,0.942}-0.004 & \cellcolor[rgb]{0.977,0.946,0.950}-0.004 & \cellcolor[rgb]{0.936,0.849,0.859}\textbf{-0.10} & \cellcolor[rgb]{0.988,0.972,0.974}-0.02 & \cellcolor[rgb]{0.948,0.877,0.885}-0.08 & \cellcolor[rgb]{0.960,0.906,0.912}\textbf{-0.006} & \cellcolor[rgb]{0.948,0.877,0.885}\textbf{-0.008} & \cellcolor[rgb]{0.970,0.981,0.975}\phantom{-}0.002 \\
\quad Cells with ${\geq}\,30$ generations & \cellcolor[rgb]{0.910,0.790,0.803}\textbf{-0.015} & \cellcolor[rgb]{0.943,0.866,0.874}-0.009 & \cellcolor[rgb]{0.968,0.924,0.929}-0.005 & \cellcolor[rgb]{0.950,0.882,0.889}\textbf{-0.08} & \cellcolor[rgb]{0.979,0.951,0.954}-0.03 & \cellcolor[rgb]{0.970,0.931,0.935}-0.04 & \cellcolor[rgb]{0.912,0.793,0.806}\textbf{-0.013} & \cellcolor[rgb]{0.927,0.829,0.840}-0.011 & \cellcolor[rgb]{0.985,0.964,0.966}-0.002 \\
\quad Without Aya and Llama & \cellcolor[rgb]{0.942,0.863,0.872}-0.010 & \cellcolor[rgb]{0.975,0.942,0.945}-0.004 & \cellcolor[rgb]{0.967,0.922,0.927}-0.006 & \cellcolor[rgb]{0.961,0.908,0.914}\textbf{-0.06} & \cellcolor[rgb]{0.999,0.997,0.997}-0.00 & \cellcolor[rgb]{0.962,0.911,0.917}-0.06 & \cellcolor[rgb]{0.936,0.849,0.859}-0.009 & \cellcolor[rgb]{0.956,0.896,0.902}-0.006 & \cellcolor[rgb]{0.980,0.953,0.956}-0.003 \\
\addlinespace
\multicolumn{10}{l}{\emph{Paraphrasing}} \\
\quad Primary specification & \cellcolor[rgb]{0.909,0.786,0.800}\textbf{-0.015} & \cellcolor[rgb]{0.920,0.813,0.825}\textbf{-0.013} & \cellcolor[rgb]{0.989,0.973,0.975}-0.002 & \cellcolor[rgb]{0.854,0.656,0.678}\textbf{-0.22} & \cellcolor[rgb]{0.982,0.959,0.961}-0.03 & \cellcolor[rgb]{0.871,0.698,0.717}-0.19 & \cellcolor[rgb]{0.915,0.800,0.813}\textbf{-0.012} & \cellcolor[rgb]{0.897,0.758,0.774}\textbf{-0.015} & \cellcolor[rgb]{0.958,0.973,0.965}\phantom{-}0.003 \\
\quad \quad + Item fixed effects & \cellcolor[rgb]{0.915,0.799,0.812}\textbf{-0.014} & \cellcolor[rgb]{0.917,0.806,0.818}\textbf{-0.014} & \cellcolor[rgb]{0.997,0.993,0.993}-0.000 & \cellcolor[rgb]{0.856,0.661,0.683}\textbf{-0.22} & \cellcolor[rgb]{0.964,0.916,0.921}-0.05 & \cellcolor[rgb]{0.892,0.745,0.762}-0.16 & \cellcolor[rgb]{0.926,0.827,0.838}\textbf{-0.011} & \cellcolor[rgb]{0.892,0.745,0.762}\textbf{-0.016} & \cellcolor[rgb]{0.917,0.946,0.930}\phantom{-}0.005 \\
\quad \quad + CoT length & \cellcolor[rgb]{0.918,0.806,0.819}\textbf{-0.014} & \cellcolor[rgb]{0.915,0.800,0.812}\textbf{-0.014} & \cellcolor[rgb]{0.993,0.996,0.994}\phantom{-}0.000 & \cellcolor[rgb]{0.856,0.661,0.683}\textbf{-0.22} & \cellcolor[rgb]{0.973,0.937,0.941}-0.04 & \cellcolor[rgb]{0.882,0.724,0.741}-0.18 & \cellcolor[rgb]{0.926,0.827,0.838}\textbf{-0.011} & \cellcolor[rgb]{0.892,0.746,0.763}\textbf{-0.016} & \cellcolor[rgb]{0.919,0.947,0.932}\phantom{-}0.005 \\
\quad \quad + Perturber similarity & \cellcolor[rgb]{0.890,0.740,0.757}\textbf{-0.018} & \cellcolor[rgb]{0.922,0.816,0.828}\textbf{-0.013} & \cellcolor[rgb]{0.968,0.924,0.929}\textbf{-0.005} & \cellcolor[rgb]{0.818,0.573,0.600}\textbf{-0.27} & \cellcolor[rgb]{0.971,0.931,0.936}-0.04 & \cellcolor[rgb]{0.847,0.641,0.664}\textbf{-0.23} & \cellcolor[rgb]{0.897,0.759,0.774}\textbf{-0.015} & \cellcolor[rgb]{0.903,0.772,0.787}\textbf{-0.014} & \cellcolor[rgb]{0.994,0.987,0.988}-0.001 \\
\quad \quad + Answer confidence & \cellcolor[rgb]{0.995,0.989,0.990}-0.001 & \cellcolor[rgb]{0.981,0.956,0.959}-0.003 & \cellcolor[rgb]{0.967,0.978,0.972}\phantom{-}0.002 & \cellcolor[rgb]{0.938,0.854,0.863}\textbf{-0.09} & \cellcolor[rgb]{0.890,0.928,0.908}\phantom{-}0.07 & \cellcolor[rgb]{0.892,0.745,0.762}-0.16 & \cellcolor[rgb]{0.915,0.945,0.929}\phantom{-}0.005 & \cellcolor[rgb]{0.994,0.986,0.987}-0.001 & \cellcolor[rgb]{0.901,0.935,0.916}\textbf{\phantom{-}0.006} \\
\addlinespace[2pt]
\quad Correct answers only & \cellcolor[rgb]{0.855,0.660,0.682}\textbf{-0.024} & \cellcolor[rgb]{0.854,0.657,0.679}\textbf{-0.024} & \cellcolor[rgb]{0.997,0.998,0.998}\phantom{-}0.000 & \cellcolor[rgb]{0.925,0.823,0.834}\textbf{-0.11} & \cellcolor[rgb]{0.902,0.771,0.785}\textbf{-0.15} & \cellcolor[rgb]{0.947,0.966,0.956}\phantom{-}0.03 & \cellcolor[rgb]{0.821,0.579,0.606}\textbf{-0.026} & \cellcolor[rgb]{0.803,0.536,0.566}\textbf{-0.029} & \cellcolor[rgb]{0.957,0.972,0.964}\phantom{-}0.003 \\
\quad Cells with ${\geq}\,30$ generations & \cellcolor[rgb]{0.900,0.765,0.780}\textbf{-0.017} & \cellcolor[rgb]{0.903,0.771,0.786}\textbf{-0.016} & \cellcolor[rgb]{0.997,0.994,0.994}-0.000 & \cellcolor[rgb]{0.922,0.817,0.829}\textbf{-0.12} & \cellcolor[rgb]{0.914,0.798,0.810}-0.13 & \cellcolor[rgb]{0.980,0.987,0.983}\phantom{-}0.01 & \cellcolor[rgb]{0.877,0.711,0.730}\textbf{-0.018} & \cellcolor[rgb]{0.885,0.731,0.748}\textbf{-0.017} & \cellcolor[rgb]{0.992,0.981,0.982}-0.001 \\
\quad Without Aya and Llama & \cellcolor[rgb]{0.924,0.821,0.833}\textbf{-0.013} & \cellcolor[rgb]{0.914,0.797,0.810}\textbf{-0.014} & \cellcolor[rgb]{0.975,0.984,0.979}\phantom{-}0.002 & \cellcolor[rgb]{0.969,0.926,0.931}-0.05 & \cellcolor[rgb]{0.951,0.886,0.893}-0.07 & \cellcolor[rgb]{0.959,0.973,0.966}\phantom{-}0.03 & \cellcolor[rgb]{0.906,0.780,0.794}\textbf{-0.014} & \cellcolor[rgb]{0.893,0.749,0.765}\textbf{-0.016} & \cellcolor[rgb]{0.969,0.980,0.974}\phantom{-}0.002 \\
\bottomrule
\end{tabular}